\documentclass[sigconf]{acmart}

\AtBeginDocument{%
  }

\setcopyright{acmlicensed}
\copyrightyear{2026}
\acmYear{2026}
\acmDOI{XXXXXXX.XXXXXXX}
\acmConference[...]{Proceedings of the nth ACM ...}{...}{...}

\acmISBN{978-1-4503-XXXX-X/2018/06}

\usepackage{amsmath,amsfonts}
\usepackage{algorithmic}
\usepackage{algorithm}
\usepackage[caption=false,font=normalsize,labelfont=sf,textfont=sf]{subfig}
\usepackage{textcomp}
\usepackage{stfloats}
\usepackage{url}
\usepackage{verbatim}
\usepackage{graphicx}
\usepackage[]{hyperref}
\usepackage{booktabs}
\usepackage{multirow}
\usepackage{wrapfig}
\usepackage{orcidlink}

\usepackage{pifont}
\newcommand{\cmark}{\ding{51}}%
\newcommand{\xmark}{\ding{55}}%

\usepackage{colortbl, xcolor}
\usepackage{paralist}

\newcommand{\tool}{\textsc{VideoLights}~}
\newcommand{\toolzs}{\textsc{VideoLights-ZS}~}
\newcommand{\toolpt}{\textsc{VideoLights-pt}~}
\newcommand{\toolb}{\textsc{VideoLights-B}~}
\newcommand{\toolbzs}{\textsc{VideoLights-B-ZS}~}
\newcommand{\toolbpt}{\textsc{VideoLights-B-pt}~}
\newcommand{\toolnospace}{\textsc{VideoLights}}
\newcommand{\toolnospacept}{\textsc{VideoLights-pt}}
\newcommand{\toolnospaceb}{\textsc{VideoLights-B}}
\newcommand{\toolnospacebpt}{\textsc{VideoLights-B-pt}}

\begin{document}

\title[Feature Refinement and Cross-Task Alignment Transformer for Joint Video Highlight Detection and Moment Retrieval]{\textbf{\textsc{VideoLights}}: Feature Refinement and Cross-Task Alignment Transformer for Joint Video Highlight Detection and Moment Retrieval}

\acmSubmissionID{265}  

\author{Dhiman Paul}
\authornote{Both authors contributed equally to this research.}
\email{dhiman.paul@northsouth.edu}
\orcid{0009-0005-1504-6911}
\affiliation{%
  \institution{North South University}
  \city{Dhaka}
  \country{Bangladesh}
}

\author{Md Rizwan Parvez}
\authornotemark[1]
\email{mparvez@hbku.edu.qa}
\orcid{0000-0002-3708-7803}
\affiliation{%
  \institution{Qatar Computing Research Institute (QCRI)}
  \country{Qatar}
}

\author{Nabeel Mohammed}
\email{nabeel.mohammed@northsouth.edu}
\orcid{0000-0002-7661-3570}

\author{Shafin Rahman}
\email{shafin.rahman@northsouth.edu}
\orcid{0000-0001-7169-0318}

\affiliation{%
  \institution{North South University}
  \city{Dhaka}
  \country{Bangladesh}
}

\renewcommand{\shortauthors}{Paul et al.}

\begin{abstract}

  Prevailing joint prediction transformers for Video Highlight Detection and Moment Retrieval (HD/MR) exhibit deficiencies in handling cross-task dynamics, achieving robust video-text alignment, and utilizing effective attention mechanisms, with the potential of Large Language/Vision-Language Models (LLMs/LVLMs) being largely untapped. This paper introduces \textbf{\toolnospace}, a novel HD/MR framework addressing these limitations by incorporating: (i) Convolutional Projection and Feature Refinement modules with an alignment loss for enhanced video-text feature congruity; (ii) a Bi-Directional Cross-Modal Fusion network for strongly coupled query-aware representations; (iii) a Uni-directional joint-task feedback mechanism for synergistic task improvement; (iv) hard positive/negative losses for adaptive learning; and (v) the leveraging of LVLMs (e.g., BLIP-2) for superior multimodal feature integration and intelligent pre-training with synthetic data. Comprehensive evaluations on QVHighlights, TVSum, and Charades-STA benchmarks demonstrate that \textbf{\tool}  significantly surpasses existing baselines, establishing new state-of-the-art performances. Codes and model checkpoints are available at \emph{TBA}.
\end{abstract}

\ccsdesc[500]{Computing methodologies~Visual content-based indexing and retrieval}
\ccsdesc[300]{Computing methodologies~Scene understanding}

\keywords{video highlight detection, moment retrieval, video grounding, feature refinement}

\received{31 July 2025}
\received[revised]{N/A}
\received[accepted]{N/A}

\maketitle

\section{Introduction}
\label{sec:intro}

The proliferation of digital devices, platforms, and internet usage has resulted in a wealth of online video content \cite{apostolidis2021video,Wu_2017}.  However, navigating this vastness presents a significant challenge, hindering users' ability to locate specific points of interest \cite{anne2017localizing,apostolidis2021video}. Hence, Video Highlight Detection (HD) and Moment Retrieval (MR, which assess video clip saliency and identify significant moments for user queries, have become essential for video analysis—streamlining content management, recommendation, creation, editing, and event detection.  Due to the shared goal of ranking/localizing relevant clips and commonalities in multi-modal models \& data, recent work has begun jointly modeling HD/MR using transfer learning \cite{lei2021detecting,Liu_2022_CVPR,yang2024taskweave, Moon_2023_CVPR,lin2023univtg,jang2023knowing,Sun_Zhou_Chen_Xie_2024,Wang_2024}.

Standard approaches for joint Moment Retrieval (MR) and Highlight Detection (HD) typically rely on extracting video and text features using pre-trained encoders such as CLIP~\cite{radford2021learning} and SlowFast~\cite{feichtenhofer2019slowfast}, projecting them into a common latent space, and fusing them for downstream processing. Depending on the architecture, these fused features are either concatenated~\cite{lei2021detecting}, used as input to a transformer with cross-attention mechanisms~\cite{Moon_2023_CVPR}, or processed through isolated modality-specific encoders with deferred interaction~\cite{Liu_2022_CVPR}. Recent models such as TaskWeave~\cite{yang2024taskweave} and TR-DETR~\cite{Sun_Zhou_Chen_Xie_2024} have further explored the reciprocal nature of MR and HD, leveraging the observation that video segments relevant to a query often exhibit high saliency.

Despite these advancements, current joint MR/HD models exhibit key limitations. First, \emph{semantic misalignment} persists due to the inadequacy of simple fusion strategies (e.g., projection or concatenation) in capturing complex intra- and inter-modal dependencies\cite{Moon_2023_CVPR}. This challenge is exacerbated by the discrepancy between brief textual queries (e.g., in QVHighlights~\cite{lei2021detecting}, Charades-STA~\cite{gao2017tall}) and lengthy, often noisy videos containing non-relevant clips, where equally weighted attention dilutes relevance. Although TR-DETR incorporates visual feature refinement, the alignment issue needs further research. Second, most models adopt \emph{uni-directional} cross-modal attention (text-to-video), neglecting the empirical benefits of bidirectional fusion~\cite{Yuan2019aaaimoment,badamdorj2021joint,xu2023mhdetr}. Third, although recent works~\cite{yang2024taskweave,Sun_Zhou_Chen_Xie_2024} begin exploring HD-MR reciprocity, most models fail to exploit their mutual reinforcement. Finally, reliance on auxiliary data such as ASR transcripts~\cite{lei2021detecting, Liu_2022_CVPR, xiao2023bridging_uvcom_cvpr_2024} or external synthetic corpora~\cite{lin2023univtg} may not always be feasible or optimal. The potential of large vision-language models (LVLMs) in addressing these challenges also remains underexploited.

To overcome these limitations, we propose \textbf{\toolnospace}, a unified framework that holistically integrates cross-modal and cross-task dynamics to advance joint video highlight detection and moment retrieval. \toolnospace addresses the aforementioned gaps through the following key components:

\begin{compactenum}
\item \textbf{Feature Refinement and Alignment (FRA) Module}: A CNN-based module that captures local and global interactions across modalities, supported by an alignment loss to bridge text-video correspondence and mitigate semantic misalignment.

\item \textbf{Bi-Directional Cross-Modal Fusion (Bi-CMF) Network}: A three-stage hierarchical attention mechanism enabling bidirectional information flow between video and text features, enhancing semantic fusion beyond traditional uni-directional schemes.

\item \textbf{Unidirectional Joint-Task Feedback Mechanism (Uni-JFM)}: Extends the idea of MR2HD~\cite{Sun_Zhou_Chen_Xie_2024} by introducing task-specific and task-coupled losses, including saliency-level cosine similarity, to reinforce reciprocal supervision across MR and HD tasks.

\item \textbf{Adaptive Error Correction}:  Incorporates hard positive and negative mining to address persistent saliency errors and enhance robustness.

\item \textbf{Intelligent Model Pre-training}: Leverages BLIP-2 to generate high-quality image-text pairs for weakly supervised pre-training, circumventing the dependency on ASR-based captions, and enhancing generalizability.
\end{compactenum}

We validate \textbf{\toolnospace} on three benchmarks: QVHighlights~\cite{lei2021detecting}, TVSum~\cite{song2015tvsum}, and Charades-STA~\cite{gao2017tall}—achieving state-of-the-art results with average gains of 4.29\%, 1.98\% and 0.7\%, respectively. Extensive ablation studies, qualitative analyses, and pre-training evaluations further demonstrate the effectiveness and scalability of our approach.

\begin{figure*}[!t]
    \centering
    \includegraphics[ clip, width=.95\textwidth]{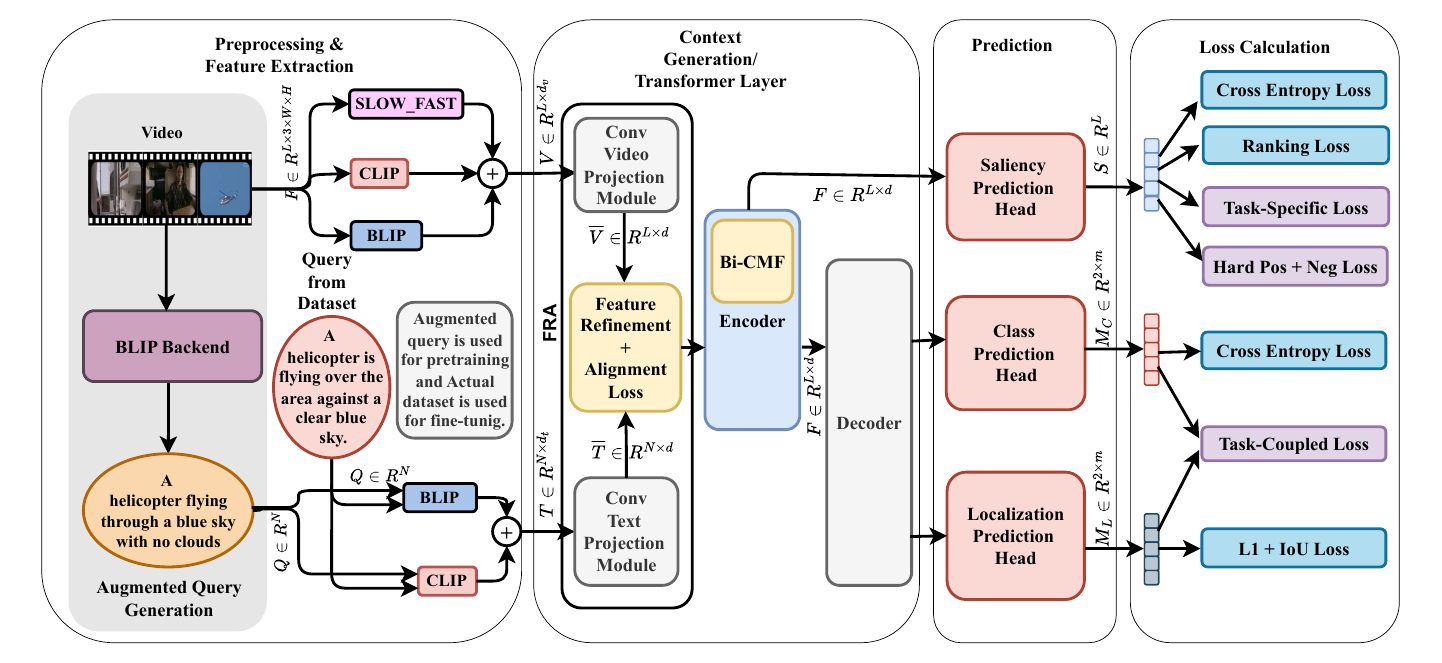}
    \vspace{-5pt}
    \caption{Overall \textbf{\toolnospace} architecture. The FRA module models video-text correlations from projected embeddings, which are then refined by the Bi-CMF encoder. A trainable saliency vector predicts output levels, while class and moment prediction heads generate logits and video moments. Cross-task feedback is provided by saliency cosine similarity and task-coupled HD/MR losses (\emph{Uni-JFM}), with new losses highlighted in purple.}
    \Description{Model structure}
    \label{fig:model_structure}
    \vspace{-5pt}
\end{figure*}

\section{Related Work}
\label{sec:related_works}

\noindent\textbf{Earlier HD/MR Transformers:}
Early HD \& MR works can be broadly categorized into two-stage approaches~\cite{anne2017localizing,hendricks2018localizing,gao2017tall,zeng2021multi,zhang2020learning,xiao2021boundary} and one-stage models~\cite{zhao2021cascaded,xiao2021natural,liu2018temporal,zhang2020learning,zhang2021multi,wang2021structured,zhang2020span,mun2020local,liu2021context,zeng2020dense}. Recently, transformer-based architectures have dominated this area following DETR \cite{carion2020end}, which eliminated anchors and non-maximum suppression. Notable contributions include Moment-DETR \cite{lei2021detecting}, which introduced the QVHighlights dataset, and UMT \cite{Liu_2022_CVPR}, which integrates multimodal (video and audio) data but sacrifices the moment decoder and bipartite matching, degrading MR performance. Other approaches, such as TVT \cite{lei2020tvr} and FVMR \cite{gao2021fast}, focus on incorporating additional modalities or improving efficiency, while R²-Tuning \cite{liu2024tuning} leverages CLIP’s multi-layer features for parameter-light temporal grounding. Recent methods like TaskWeave \cite{yang2024taskweave} and TR-DETR \cite{Sun_Zhou_Chen_Xie_2024} explore cross-task dependencies, yet our work specifically addresses both cross-modal and cross-task interactions in a unified HD/MR framework. Text-video relevance has been explored utilizing dummy tokens in CG-DETR~\cite{moon2024correlationguided_cgdetr}, while the effectiveness of Multimodal Large Language Models like BLIP has also been investigated in Mr. Blip~\cite{meinardus2024surprisingeffectivenessmultimodallarge}. SGDETR~\cite{gordeev2024saliencyguideddetrmomentretrieval}, FlashVTG~\cite{cao2024flashvtgfeaturelayeringadaptive}, and InternVideo2~\cite{wang2024internvideo2scalingfoundationmodels} along with model improvements utilized InternVideo2 visual features for their work.

\noindent\textbf{Cross-modal Learning:} 
Cross-modal learning integrates visual and textual modalities for richer semantic understanding. Prior works like TERAN~\cite{messina2021fine}, HGSPN~\cite{hu2019hierarchical}, AVS~\cite{morgado2020learning}, and \cite{badamdorj2021joint} explore various fusion strategies. ABLR~\cite{Yuan2019aaaimoment} uses bidirectional attention but is limited to moment retrieval (MR). UnLoc~\cite{Yan_2023_ICCV} unifies tasks via CNN-based pyramids and CLIP embeddings. In contrast, we propose a three-stage sequential process for joint MR/HD tasks that hierarchically refines video and text representations through mutual attention, enabling more targeted learning of complex cross-modal relationships. This is further enhanced via cross-task supervision within a unified MR/HD framework.

\noindent\textbf{Weakly Supervised Training:} Recent studies have demonstrated that weakly supervised pretraining, often using ASR-generated captions \cite{lei2021detecting,xiao2023bridging_uvcom_cvpr_2024,Liu_2022_CVPR}, improves model performance. For instance, \cite{Yan_2023_ICCV} pre-trains their CLIP backend with Kinetics-700 \cite{carreira2022short} before fine-tuning on downstream tasks, while UniVTG \cite{lin2023univtg} leverages large corpora from Ego4D \cite{grauman2022ego4d} and VideoCC \cite{nagrani2022learning}. In contrast, our method demonstrates robustness without relying on such extensive data diversity. Additionally, work in text-only contexts \cite{parvez-etal-2023-retrieval} suggests that combining different encoders can enhance supervision.

\section{Proposed \textbf{\toolnospace} Model}

We present \textbf{\toolnospace}, our joint prediction HD/MR model that enables learning from cross-modal (text vs video) and cross-task (HD vs MR) interplays. \textbf{\tool} features a unique composite of a Feature Refinement and Alignment Network, a Bi-Directional Cross-Modal Fusion Network, a Unidirectional Join-Task Feedback module, advanced appetite loss functions, and intelligent  pre-training. 
\textbf{\tool} pipleline is shown in Figure~\ref{fig:model_structure}. 

\subsection{Model Overview}

Highlight Detection (HD) and Moment Retrieval (MR) aim to estimate the saliency of video clips and identify significant moments for a given text query. Given a video of \( L \) clips, we define the video clips as \( F \in \mathbb{R}^{L \times 3 \times W \times H} \), where \( W \) and \( H \) denote the width and height of the video, and \( 3 \) represents the number of color channels. The feature representation of the video is denoted as \( V \in \mathbb{R}^{L \times d_v} \), where \( d_v \) is the feature dimension extracted by a frozen video encoder. Given a text query of \( N \) tokens, the representation of the text is denoted as \( T \in \mathbb{R}^{N \times d_t} \), where \( d_t \) is the feature dimension extracted by a frozen text encoder. With these representations and given the video and the text, our goal is twofold: for Moment Retrieval (MR), we aim to determine all the moments \( M \in \mathbb{R}^{2 \times m} \), where each moment consists of a central coordinate \( m_c \) and width \( m_\sigma \), identifying \( m \) such moments within the video. For Highlight Detection (HD), we aim to rank the saliency scores \( S \in \mathbb{R}^L \) for each clip in the video to detect highlights.

\noindent\textbf{Embeddings:} We compute the initial feature sets \( V \) and \( T \) from multiple different VLPs as follows:
\(
  T = \text{clip}(Q) \oplus \text{blip}(Q)  
  \) and \(
  V = \text{clip}(F) \oplus \text{slowfast}(F) \oplus \text{blip}(F) 
  \)

Here $\oplus$ operator denotes concatenation of the features and \(\text{clip}\), \(\text{blip}\), and \(\text{slowfast}\) refer to frozen CLIP ~\cite{radford2021learning}, BLIP-2 ~\cite{li2023blip}, and Slow-Fast models ~\cite{feichtenhofer2019slowfast} respectively. 

\noindent\textbf{Projection and Alignment:} 
To resolve dimensional mismatch between the representations of video \( V \) and text \( T \), we apply a feed forward convolutional network (FFCNN) for the alignment of local features, yielding \( \overline{V} \in \mathbb{R}^{L \times d} \) and \( \overline{T} \in \mathbb{R}^{N \times d} \), respectively, where \( d \) is the dimension of shared features (see Section~\ref{sec:fra}).

Then, both video and text representations are fed into the video-query refinement module to learn query-attended video representations and highlight relevant tokens (see Section~\ref{sec:fra}).

\noindent\textbf{Encoder with Cross-Modal Interaction:} 
Refined video and query tokens are processed via the \emph{Bi-CMF} module (see Section \ref{sec:bi-cmf}), which fuses features to capture inter-relevance, forming a query-injected video representation. A multilayer encoder applies self-attention to this fused representation prior to saliency prediction.

\noindent\textbf{Decoder with Cross-Task Dynamics:} 
The encoder output is passed to a decoder following \cite{Moon_2023_CVPR}. This output informs class and localization prediction heads. Negative relations between irrelevant video-query pairs refine the response. We introduce \emph{Uni-JFM}, a unidirectional cross-task feedback network that computes task-specific and cross-task losses (see Section \ref{sec:uni-jfm}).

\noindent\noindent\textbf{Adaptive Learning and Loss Functions} 
\textbf{\toolnospace} employs distinct losses for moment retrieval and highlight identification. For moment retrieval, we use L1, gIoU~\cite{union2019metric} loss $\mathcal{L}_\text{gIoU}(m, \overline{m})$ where $m$ and $\overline{m}$ are predicted and ground truth moments, and cross-entropy loss $\mathcal{L}_\text{cls}$ as in~\cite{lei2021detecting}. For highlight identification, we apply margin ranking loss $\mathcal{L}_\text{rank}$, rank contrastive loss $\mathcal{L}_\text{cont}$~\cite{Moon_2023_CVPR}, and entropy loss. In addition, we incorporate alignment loss $\mathcal{L}_{\text{align}}$ from FRA, $\mathcal{L}_\text{Uni-JFM}$ from Section~\ref{sec:uni-jfm} and adaptive hard negative and positive loss, $\mathcal{L}_\text{hdl}$ (see Section~\ref{loss-functions}), to penalize persistent saliency errors.

The moment loss is formulated as:
\begin{equation*}
\mathcal{L}_\text{mr} = \lambda_{L1}\|m-\overline{m}\| + \lambda_\text{gIoU}\mathcal{L}_\text{gIoU}(m, \overline{m}) + \lambda_\text{cls}\mathcal{L}_\text{cls}.
\end{equation*}

The overall saliency loss is defined by:
\begin{align*}
\mathcal{L}_{hl} =& \lambda_\text{rank}\mathcal{L}_\text{rank} + \lambda_\text{cont}\mathcal{L}_\text{cont} 
                  + \lambda_\text{hdl}\mathcal{L}_\text{hdl} + \mathcal{L}_\text{Uni-JFM}.
\end{align*}

Incorporating the alignment loss $\mathcal{L}_{\text{align}}$ (see Section~\ref{sec:fra}), the final total loss becomes:
\begin{equation*}
\mathcal{L}_\text{total} = \lambda_\text{sal}\mathcal{L}_\text{hl} + \mathcal{L}_\text{mr} + \lambda_\text{al}\mathcal{L}_{\text{align}}
\end{equation*}
where $\lambda_\text{hdl}$, $\lambda_\text{sal}$ and $\lambda_\text{al}$ balance the contributions. In the following, we discuss \emph{FRA}, \(\mathcal{L}_{\text{align}}\), \emph{Bi-CMF} and \emph{Uni-JFM} modules, the adaptive losses $\mathcal{L}_{\text{hard}_\text{neg}}$, $\mathcal{L}_{\text{hard}_\text{pos}}$, and our pretraining procedure.

\subsection{Feature Refinement \& Alignment Network}
\label{sec:fra}

\begin{figure*}[t]
    \centering
    \includegraphics[trim=10mm 3mm 5mm 0mm, clip, width=\textwidth]{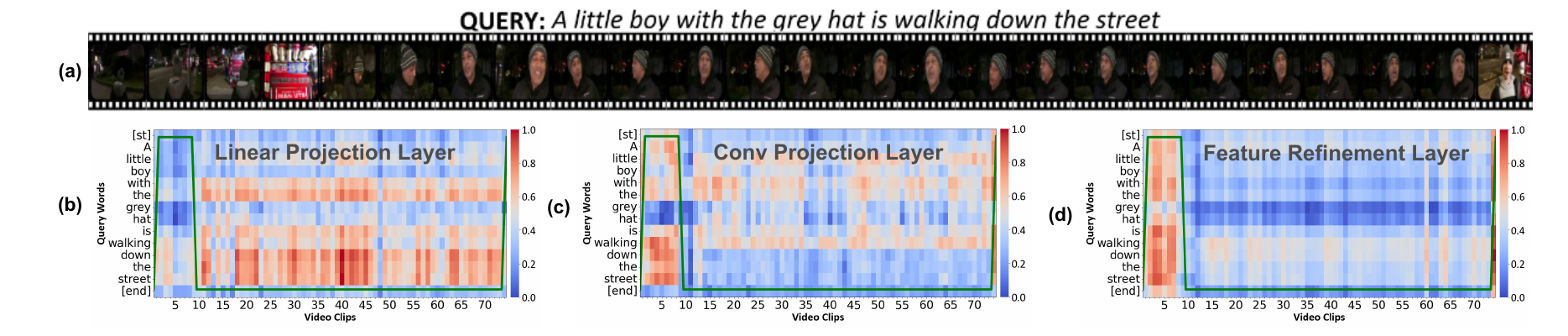}
    \vspace{-16pt}
    \caption{(a) is the input video, (b) and (c) are correspondence maps of query and video tokens using linear and convolution layers, respectively, which show that queries are more aligned for the convolution layer, video, and text than linear projection layers. (d) The effect of the Feature Refinement module that effectively aligns video and text tokens that match ground truth saliency levels (green line) in each heat map saliency level is shown with \textcolor{teal}{green} line plot.}
    \Description{FRA effect}
    \vspace{-7pt}
    \label{fig:fra}
\end{figure*}


Text queries are typically concise and informative, whereas videos often include irrelevant segments. Standard self- and cross-attention mechanisms uniformly weight all video tokens, diluting focus on salient regions. To mitigate this, we introduce the \textbf{Feature Refinement and Alignment (FRA) Network}, which enhances both local (token-level) and global (video-sentence) alignment by emphasizing query-relevant video tokens through a two-stage process.

In \textbf{Stage 1}, a convolutional projection layer captures local representations and aligns video-text features by adjusting token dimensions using a multilayer 1D Feed-Forward Convolutional Network (FFCNN) with ReLU activation. As shown in Table~\ref{tab:effect_projection_type} in Appendix, FFCNN outperforms standard linear projection, with qualitative improvements illustrated in Figure~\ref{fig:fra}.

This projection transforms \( V \in \mathbb{R}^{L \times d_v} \) into \( \overline{V} \in \mathbb{R}^{L \times d} \) and \( T \in \mathbb{R}^{N \times d_t} \) into \( \overline{T} \in \mathbb{R}^{N \times d} \):
\begin{gather*}
     \overline{V} = \text{relu} (\text{FFCNN} (V)),  \qquad
     \overline{T} = \text{relu} (\text{FFCNN} (T)) 
\end{gather*}

In \textbf{Stage 2}, a feature refinement layer enhances global alignment and highlights query-relevant video tokens by computing a correspondence map between locally aligned video and text tokens, extracting sentence-level features, generating a similarity matrix with video tokens, and aggregating results by concatenating and projecting to the hidden layer dimension \( d \) using a 1D convolutional network. Sentence-level alignment is crucial in this process, as it captures the global semantic correspondence between the entire query and the video context, allowing the model to attend to relevant segments beyond token-level associations. This refinement enables the model to emphasize semantically meaningful regions, and subsequent attentions are applied to these highlighted tokens, resulting in more precise and context-aware modality fusion.

The refinement process is formulated as:
\begin{align*}
V_Q=& \overline{V} \cdot \overline{T}^T , \qquad
S= \text{pool}  (\overline{T}),  \qquad  
V_S= \overline{V} \cdot S^T ,  \qquad \\
S_v=& S \cdot 1_{1\times V \times 1},  \qquad
V_r= \text{conv}( \overline{V} \oplus V_Q\oplus V_S \oplus S_v) &
\end{align*}
where \( \cdot \) is matrix multiplication and \( V_r \) the refined video tokens.

To achieve this refinement, we employ a video-text alignment loss. The Video-text alignment loss ($\mathcal{L}_{\text{align}}$) is computed by first estimating the saliency of refined video tokens with respect to query tokens, followed by matching these estimates to ground-truth saliency scores. The alignment loss is defined as:
\begin{equation*}
\mathcal{L}_{\text{align}} = \frac{1}{B} \sum_{b=1}^{B} \left( 1 - 
\frac{\text{norm}(\mathbf{s}_b) \cdot \text{norm}(\hat{\mathbf{s}}_b)}{\lVert \text{norm}(\mathbf{s}_b) \rVert \lVert \text{norm}(\hat{\mathbf{s}}_b) \rVert}
\right)
\end{equation*}
where $b \in B$ indexes the mini-batch, $\text{norm}(\cdot)$ denotes L2 normalization, $\mathbf{s}_b$ is the ground-truth saliency, and $\hat{\mathbf{s}}_b = \text{sim}(\overline{T}, V_r)$ is the predicted saliency via cosine similarity between query tokens $\overline{T}$ and refined video features $V_r$.

Figure~\ref{fig:fra} contrasts the standard linear projection with the convolutional approach of FRA, showing its improved focus on relevant tokens, which enhances similarity scores aligned with the saliency of the ground truth.

\subsection{Bi-Directional Cross-Modal Fusion Network}
\label{sec:bi-cmf}

To foster strongly coupled, query-oriented video representations and achieve text-video semantic disambiguation, we introduce  Bi-Directional Cross-Modal Fusion Network (\emph{Bi-CMF}), which leverages bidirectional cross-attention, a technique notably underexplored in joint MR/HD tasks.
It features three multi-head attention layers for cross-attention. Initially, a cross-attention layer uses projected video features as queries, while text data with positional embedding serve as keys and values, identifying video tokens conditioned by textual tokens.
Similarly, another cross-attention layer is utilized to discern projected textual tokens (query) features conditioned by video tokens fused with positional embedding (keys and values), enabling the identification of textual features pertinent to the video. 
Subsequently, conditioned video tokens are used as queries, while conditioned textual tokens serve as keys and values in the final cross-attention layer, yielding fused contextual information that emphasizes video tokens relevant to the query.
\begin{gather*}
    V_{T} = \text{attn}(V_r, \overline{T}, \overline{T}),    \qquad
    T_{V} = \text{attn}(\overline{T}, V_r, V_r),  \qquad \\
    V_\text{attn} = \text{attn}(\overline{V}_{T}, \overline{T}_{V}, \overline{T}_{V})
\end{gather*}

\begin{figure}[!t]
    \centering
    \includegraphics[trim = 15mm 6.7mm 11mm 4.5mm, clip,width=.5\textwidth]{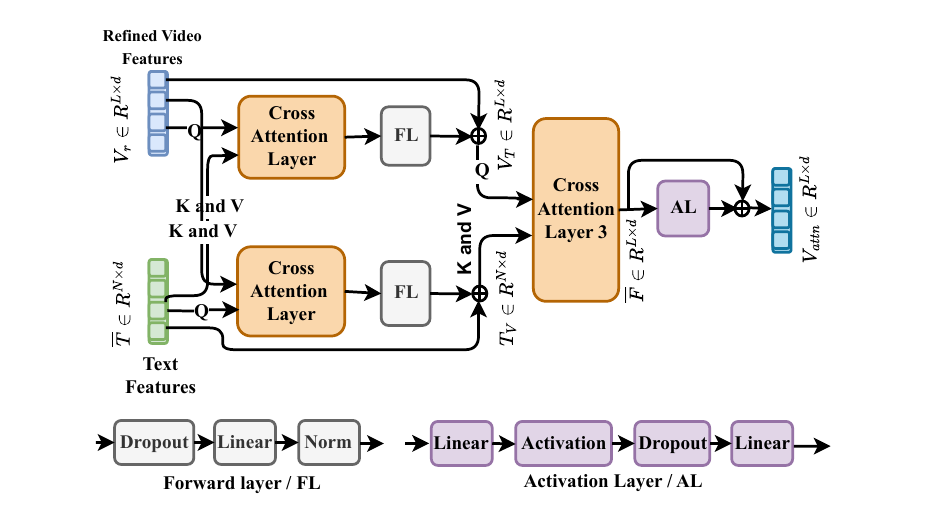}
    \vspace{-15pt}
    \caption{Bi-CMF Module. It learns query-oriented video via text2video, video2text, then text2video attentions. In this process, dropout and normalization are applied after each step, and activation is applied at the last stage.
    } 
    \Description{Bi-CMF Structure}
    \label{fig:bi-cnf}
    \vspace{-15pt}
\end{figure}

Residual connections~\cite{he2016deep}, layer norms~\cite{ba2016layer}, \& dropout~\cite{srivastava2014dropout} mechanisms are implemented at each stage to enhance the robustness of the model, and encodings of learnable positions are incorporated into the input of each attention layer. 
\emph{Bi-CMF} is depicted in Figure~\ref{fig:bi-cnf}. 

\subsection{Adaptive Loss Functions}
\label{loss-functions}
We aim to enhance learning by identifying and rectifying persistent model errors. To achieve this, we design novel adaptive loss functions, specifically targeting hard positives and hard negatives. For the hard negative loss, we minimize the number of predictions in the negative regions where there are no relevant clips. Given the saliency score \(\bar{S}_i\) and the ground truth saliency score \(\mathcal{S}_i\) for non-relevant clips \(i \in V_{neg}\), we define the loss,
 \(
\mathcal{L}_{\text{hard}_\text{neg}} = W_j \Sigma_{i \in V_\text{neg}} \text{abs}(\mathcal{S}_i - \bar{S}_i) 
\),
where \( W_j \) is a function of the \( j \)th epoch (for simplicity, we used \( W_j=j+1 \) that penalizes more with a higher number of epochs. As in general, \(\mathcal{S}_i\) for \(i \in V_\text{neg}\) is zero, the loss can be defined as:
\(
\mathcal{L}_{\text{hard}_\text{neg}} = W_j \Sigma_{i \in V_\text{neg}} \text{abs}(\bar{S}_i) 
\).

For hard positive cases, we use Mean Square Error, and similarly, we define the loss as:
\(
\mathcal{L}_{\text{hard}_\text{pos}} = W_j \Sigma_{i \in V_\text{pos}} \text{MSE}(\mathcal{S}_i, \bar{S}_i) 
\).
Then the total hard negative and positive loss becomes: 
\[\mathcal{L}_\text{hdl}=\mathcal{L}_{\text{hard}_\text{pos}}+\mathcal{L}_{\text{hard}_\text{neg}}\]

\subsection{Unidirection Joint-Task Feedback Module}
\label{sec:uni-jfm}
To leverage the synergies between tasks while jointly predicting HD/MR, we devise a unidirectional joint-task feedback mechanism that is a combination of a task-specific and a task-coupled loss. The task-specific loss directly optimizes the HD scores, while the task-coupled loss facilitates indirect supervision of MR by leveraging the learned representations from the HD task. We take HD as a reference task and compute its task-specific loss $\mathcal{L}_\text{ts}$. To do so, we calculate the saliency cosine similarity loss from the predicted saliency level. For saliency score $\bar{S}$ and ground truth saliency score $\mathcal{S}$ the task-specific loss $\mathcal{L}_\text{ts}$ can be defined as: 
\( \mathcal{L}_\text{ts} = 1 - \frac{\bar{S}.\mathcal{S}}{\lVert \bar{S} \rVert \lVert \mathcal{S} \rVert} \).

Next, for the task-coupled loss $\mathcal{L}_\text{tc}$, first, we use the feature vectors for MR,  \(M\) to calculate saliency scores $\bar{S}_\text{mr}$ following the MR2HD technique of ~\cite{Sun_Zhou_Chen_Xie_2024} using a GRU unit. Then, differently, we calculate the similarity between the ground truth saliency  $\mathcal{S}$ and this calculated saliency $\bar{S}_\text{mr}$. This similarity score is used as the loss function $\mathcal{L}_\text{tc} $, where 
\( 
\mathcal{L}_\text{tc} = 1 - \frac{\bar{S}_\text{mr}.\mathcal{S}}{\lVert \bar{S}_\text{mr} \rVert \lVert \mathcal{S} \rVert} 
\).

The total loss for the module becomes,
\[
\mathcal{L}_\text{Uni-JFM} = \lambda_\text{ts}\mathcal{L}_\text{ts} + \lambda_\text{tc}\mathcal{L}_\text{tc}
\]
where $\lambda_\text{ts}$ and $\lambda_\text{tc}$ hyperparameters balance $\mathcal{L}_\text{ts}$ and $\mathcal{L}_\text{tc}$.

Here, Cosine similarity is employed over cross-entropy loss to capture directional alignment between predicted and ground-truth saliency distributions, proving more effective under sparse supervision by prioritizing vector orientation over magnitude in high-dimensional spaces \cite{you2025semanticsanglecosinesimilarity, yu2020structureconsistentweaklysupervisedsalient}.

While both TR-DETR and \textbf{\tool} leverage MR–HD reciprocity, their strategies diverge. TR-DETR adopts explicit dual-task cooperation (HD2MR and MR2HD), whereas \tool introduces a lightweight Uni-JFM module that treats HD as the supervisory anchor, guiding MR through task-specific and task-coupled saliency-based losses. This design facilitates stable joint training with minimal overhead while preserving cross-task synergy.

\subsection{Pretraining}
\label{sec: pretrain}

We propose a novel multistep methodology to enhance attention-based networks' performance by addressing limitations in ASR caption-based weakly supervised training \cite{lei2021detecting,xiao2023bridging_uvcom_cvpr_2024}. ASR may not always align with or describe the content of the video of that timeframe. Our approach segments videos into 10-second intervals, generates descriptive captions using the BLIP model for representative frames, and creates synthetic data pairs from QVHighlights and Charades-STA datasets. Saliency scores are calculated based on frame-query similarity, and the resulting caption-query pairs are used for model training. Although this process may generate noisy pretrain data, the subsequent fine-tuning helps filter out irrelevant information, leading to improved generalization \cite{wu-etal-2022-noisytune} (see Appendix Section~\ref{sec:bias_filtering}). Detailed data statistics and steps are provided in Table \ref{tab:datasets} and Algorithm~\ref{alg:data_gen} in Appendix Section~\ref{sec:appendix}. 

\section{Experiments}

\begin{table*}[!t]
\caption{Results on QVHighlights \textbf{test} split. \(\dag\) represents the use of audio modality. Here, bold represents the best result, and underline represents the 2nd best result.}
\small
\label{tab:results-qvhighlights}
\vspace{-5pt}
\centering
\begin{tabular}{@{}lccccccc@{}}
\toprule
\multirow{3}{*}{Method} & \multicolumn{5}{c}{MR} & \multicolumn{2}{c}{HD} \\ \cmidrule(lr){2-8}
  & \multicolumn{2}{c}{R1} & \multicolumn{3}{c}{mAP} &  \multicolumn{2}{c}{\textgreater{}=Very Good} \\
  & @0.5 & \textit{@0.7} & @0.5 & @0.75 & Avg & mAP & HIT@1 \\ \midrule
    Moment-DETR (NIPS, 2021)~\cite{lei2021detecting} & 52.89 & 33.02 & 54.82 & 29.4 & 30.73 & 35.69 & 55.6 \\
    
    UMT (CVPR, 2022)~\cite{Liu_2022_CVPR} \(\dag\) & 56.23 & 41.18 & 53.83 & 37.01 & 36.12 & 38.18 & 59.99 \\
    MH-DETR (IJCNN, 2024)~\cite{xu2023mhdetr} & 60.05 & 42.48 & 60.75 & 38.13 & 38.38 & 38.22 & 60.51 \\
    EaTR (ICCV, 2023)~\cite{jang2023knowing} & 61.36 & 45.79 & 61.86 & 41.91 & 41.74 & 37.15 & 58.65 \\
    QD-DETR (CVPR, 2023)~\cite{Moon_2023_CVPR} & 62.40 & 44.98 & 63.17 & 42.05 & 41.44 & 39.13 & 63.1 \\ 
    UVCOM (CVPR, 2024)~\cite{xiao2023bridging_uvcom_cvpr_2024} & \underline{63.55} & 47.47 & 63.37 & 42.67 & \underline{43.18} & 39.74 & \underline{64.20}\\ 
    TR-DETR (AAAI, 2024)~\cite{Sun_Zhou_Chen_Xie_2024}  & \textbf{64.66} & \textbf{48.96} & \textbf{63.98} & \textbf{43.73} & 42.62 & \underline{39.91} & 63.42\\ 
    UniVTG (ICCV, 2023)~\cite{lin2023univtg} & 58.86 & 40.86 & 57.60 & 35.59 & 35.47 & 38.20 & 60.96  \\
    \rowcolor[gray]{.8}
    \textbf{\toolnospace} & 63.36 & \underline{48.70} & \underline{63.81} & \underline{42.87} & \textbf{43.38} & \textbf{40.57} & \textbf{65.30} \\ 
    \midrule
    Moment-DETR(pt) (NIPS, 2021)~\cite{lei2021detecting} & 59.78 & 40.33 & 60.51 & 35.36 & 36.14 & 37.43 & 60.17 \\
    UMT(pt) (CVPR, 2022)~\cite{Liu_2022_CVPR} & 60.83 & 43.26 & 57.33 & 39.12 &  38.08 & 39.12 & 62.39 \\
    QD-DETR(pt) (CVPR, 2023)~\cite{Moon_2023_CVPR} & 64.10 & 46.10 &  64.30 & 40.50 & 40.62 & 38.52 & 62.27 \\
    UVCOM(pt) (CVPR, 2024)~\cite{xiao2023bridging_uvcom_cvpr_2024} & 64.53 & 48.31 & \underline{64.78} & 43.65 & \underline{43.80} & 39.98 & 65.58 \\ 
    UniVTG(pt) (ICCV, 2023)~\cite{lin2023univtg} & \underline{65.43} & \underline{50.06} & 64.06 & \underline{45.02} & 43.63 & \underline{40.54} & \textbf{66.28} \\
    \rowcolor[gray]{.8}
    \textbf{\toolnospacept} & \textbf{68.48} & \textbf{52.53} & \textbf{67.31} & \textbf{46.76} & \textbf{45.01} & \textbf{41.48} & \underline{65.89} \\ 
    \midrule
    \rowcolor[gray]{.8}
    \textbf{\toolnospaceb} & \textbf{68.29} & \textbf{52.79} & \textbf{67.58} & \textbf{47.30} & \textbf{46.53} & \textbf{42.43} & \textbf{68.94} \\ 
    \rowcolor[gray]{.8}
    \textbf{\toolnospacebpt} & \textbf{70.36} & \textbf{55.25} & \textbf{69.53} & \textbf{49.17} & \textbf{47.94} & \textbf{42.84} & \textbf{70.56} \\ \bottomrule
\end{tabular}
\vspace{-5pt}
\end{table*}

\noindent\textbf{Datasets:} 
We evaluate \textbf{\toolnospace} using three widely recognized benchmarks to ensure a comprehensive and rigorous assessment. First, the \emph{QVHighlights} dataset~\cite{lei2021detecting} uniquely combines Moment and Highlight Detection tasks, providing extensive video annotations and maintaining evaluation impartiality through its online server. This dataset includes 12,562 YouTube videos and 10,310 annotations, with standardized data splits as per established works. Additionally, we use the \emph{Charades-STA}~\cite{gao2017tall} dataset for Moment Retrieval (MR) and the \emph{TVSum}~\cite{song2015tvsum} dataset for Highlight Detection (HD). TVSum, encompasses ten categories with five videos each.   We follow the data splits in ~\cite{Liu_2022_CVPR, xu2023mhdetr, Moon_2023_CVPR}, that consider 80\% of the dataset for training and 20\% for testing. Charades-STA, features 9,848 videos and 16,128 query texts,  We adopt the data splits in prior work QD-DETR~\cite{Moon_2023_CVPR} with 12,408 samples for training and 3,720 for testing. Our adherence to these standardized splits and the diversity of datasets underscore our commitment to a robust and fair evaluation of \textbf{\toolnospace}.

\noindent\textbf{Evaluation Metrics:} We follow the established evaluation metric standards from ~\cite{lei2021detecting,Liu_2022_CVPR,Moon_2023_CVPR,xu2023mhdetr,jang2023knowing}. For moment retrieval, we calculate Recall@1 with predetermined thresholds of 0.5 and 0.7, mean average precision (mAP) with Intersection over Union (IoU) thresholds of 0.5 and 0.75, and average mAP across multiple IoU thresholds that range from 0.50 to 0.95. The same standards are applied to the QVHighlights dataset. For highlight identification, our evaluations include measuring mAP and HIT@1, indicating the hit ratio for the clip with the highest score.

\noindent\textbf{Implementation details:}\label{sec:implementation_details} We train four main models per dataset: \textbf{\toolnospace} and its pre-trained version \textbf{\toolnospacept} (utilizing CLIP and SlowFast features), alongside \textbf{\toolnospaceb} and \textbf{\toolnospacebpt} (incorporating CLIP, BLIP, and SlowFast features); \textbf{-PT} denotes pre-training on synthetic data. For TVSum, a \textbf{\toolnospace} variant employs I3D visual features~\cite{carreira2017quo} (pre-trained on Kinetics 400~\cite{kay2017kinetics}) for TR-DETR~\cite{Sun_Zhou_Chen_Xie_2024} comparable evaluation. The models are configured with a hidden size of $d = 256$, one Bi-CMF layer, three encoder and decoder layers, and 10 moment queries. Dropout rates are 0.1 for transformer layers and 0.5 for input projection layers~\cite{lei2021detecting}. Loss weights are set as $\lambda_{\text{L1}}=10$, $\lambda_{\text{gIoU}}=1$, $\lambda_{\text{cls}}=4$, $\lambda_{\text{sal}}=1$, $\lambda_{\text{rank}}=1$, $\lambda_{\text{cont}}=1$, and $\Delta=0.2$. The models use Xavier initialization~\cite{glorot2010understanding} and are optimized with AdamW~\cite{loshchilov2017decoupled} using a learning rate of $10^{-4}$ and a weight decay of $10^{-4}$. Training runs for 200 epochs. Batch sizes and learning rates are set to (32, $10^{-4}$) for Charades-STA and (4, $10^{-3}$) for TVSum. All experiments use T4 and RTX 3050 Ti GPUs. \textbf{\tool} and \textbf{\toolb} have 10.4M/10.8M parameters, use 1.84 GB of GPU memory, and achieve the best results.  The full hyperparameter settings are given in the Appendix~\ref{sec:reproducibility_statement}, Table~\ref{tab:exp_details}.


\begin{table*}[!t]
\centering
\caption{
Evaluation of highlight detection methods on TVSum using Top-5 mAP. \(\dag\) represents the use of audio modality. \(\ddag\) indicates the use of I3D for visual features. Here, bold represents the best result, and underline represents the 2nd best result.}
\small
\label{tab:results-tvsum}
\vspace{-5pt}
\begin{tabular}{@{}lcccccccccccc@{}}
\toprule
\multicolumn{1}{l}{Methods} & VT & VU & GA & MS & PK & PR & FM & BK & BT & DS & Avg. \\ \midrule
sLSTM (ECCV, 2016)~\cite{zhang2016video}\(\ddag\) & 41.1 & 46.2 & 46.3 & 47.7 & 44.8 & 46.1 & 45.2 & 40.6 & 47.1 & 45.5 & 45.1 \\
SG (CVPR, 2017)~\cite{mahasseni2017unsupervised}\(\ddag\) & 42.3 & 47.2 & 47.5 & 48.9 & 45.6 & 47.3 & 46.4 & 41.7 & 48.3 & 46.6 & 46.2 \\
LIM-S (CVPR, 2019)~\cite{xiong2019less}\(\ddag\) & 55.9 & 42.9 & 61.2 & 54.0 & 60.3 & 47.5 & 43.2 & 66.3 & 69.1 & 62.6 & 56.3 \\
Trailer (ECCV, 2020)~\cite{wang2020learning}\(\ddag\) & 61.3 & 54.6 & 65.7 & 60.8 & 59.1 & 70.1 & 58.2 & 64.7 & 65.6 & 68.1 & 62.8 \\
SL-Module (ICCV, 2021)~\cite{xu2021cross}\(\ddag\) & 86.5 & 68.7 & 74.9 & 86.2 & 79 & 63.2 & 58.9 & 72.6 & 78.9 & 64.0 & 73.3 \\
UMT (CVPR, 2022)~\cite{Liu_2022_CVPR}\(\dag\)\(\ddag\) & 87.5 & 81.5 & 81.5 & 81.5 & 81.4 & 87.0 & 76.0 & 86.9 & 84.4 & 79.6 & 83.1 \\
MH-DETR (IJCNN, 2024)~\cite{xu2023mhdetr} & 86.1 & 79.4 & 84.3 & 85.8 & 81.2 & 83.9 & 74.3 & 82.7 & 86.5 & 71.6 & 81.6 \\
QD-DETR (CVPR, 2023)~\cite{Moon_2023_CVPR}\(\ddag\) & 88.2 & 87.4 & 85.6 & 85.0 & 85.8 & 86.9 & 76.4 & 91.3 & \underline{89.2} & 73.7 & 85.0 \\
UVCOM (CVPR, 2024)~\cite{xiao2023bridging_uvcom_cvpr_2024}\(\ddag\) & 87.6 & 91.6 & 91.4 & \underline{86.7} & \underline{86.9} & 86.9 & 76.9 & 92.3 & 87.4 & 75.6 & 86.3 \\


 TR-DETR (AAAI, 2024)~\cite{Sun_Zhou_Chen_Xie_2024}\(\ddag\) & \underline{89.3} & \textbf{93.0} & \underline{94.3} & 85.1 & \textbf{88.0} & \underline{88.6} & \textbf{80.4} & 91.3 & \textbf{89.5} & \textbf{81.6} & \textbf{88.1} \\

\rowcolor[gray]{.8}
\textbf{\toolnospace} \(\ddag\) & \textbf{89.8} & \underline{88.7}  & \textbf{95.0} & \textbf{88.0} & 83.6 & \textbf{90.1} & \underline{79.4} & \textbf{94.2} & 88.6 & \underline{81.2} & \underline{87.9} \\

\midrule
UniVTG (ICCV, 2023)~\cite{lin2023univtg} & \underline{83.9} & \underline{85.1} & \underline{89.0} &  \underline{80.1} &  \underline{84.6} &  \underline{81.4} &  \underline{70.9} &  \underline{91.7} &  \underline{73.5} &  \underline{69.3}  &  \underline{81.0} \\
\rowcolor[gray]{.8}
\textbf{\toolnospace}  & \textbf{89.1} & \textbf{92.7} & \textbf{92.3} & \textbf{86.7} & \textbf{89.8} & \textbf{88.9} & \textbf{78.5} & \textbf{94.0} & \textbf{87.4} & \textbf{78.3} & \textbf{87.8} \\ 
\midrule
UniVTG (pt) (ICCV, 2023)~\cite{lin2023univtg} & \textbf{92.0} & \underline{77.8} &  \underline{89.8} &  \underline{83.8} &  \underline{82.2} &  \underline{85.8} &  \underline{74.3} &  \underline{91.8} &   \textbf{90.5} &  \underline{77.6} & \underline{84.6} \\
\rowcolor[gray]{.8}
\textbf{\toolnospacept} & \underline{90.8} & \textbf{91.8} & \textbf{95.0} & \textbf{85.3} & \textbf{88.6} & \textbf{89.6} & \textbf{76.7} & \textbf{94.0} & \underline{88.5} & \textbf{78.6} & \textbf{87.9} \\ 
 \midrule
 \rowcolor[gray]{.8}
\textbf{\toolnospaceb} & 91.3 & 92.5 & 93.3 & 84.3 &  88.0 &  88.3 & 77.3  & 92.7  & 88.2 & 81.6 & 87.75 \\
\rowcolor[gray]{.8}
\textbf{\toolnospacebpt} & 91.4 & 88.2 & 93.0 & 95.2 & 87.2 & 89.1 & 76.1 & 95.1 & 88.6 & 81.3 & 88.52 \\ 
 \bottomrule
\end{tabular}
\vspace{-5pt}
\end{table*}

\subsection{Main Results}
\label{sec:main-results}

\noindent \textbf{Performance in QVHighlights:}
Table~\ref{tab:results-qvhighlights} compares various methods on the QVHighlights test split for both moment retrieval (MR) and highlight detection (HD) tasks. Our framework, \textbf{\tool}, achieves state-of-the-art results across most metrics. While TR-DETR shows marginal MR gains (~0.33\%), \textbf{\tool}'s distinct architecture (FRA, Bi-CMF, Uni-JFM) excels in mAP@Avg and HD (>1.5\%), with its FRA module also proven to enhance TR-DETR's performance (Table~\ref{tab:fra_on_diff_methods}). In MR, the \textbf{\toolnospacebpt} model obtains the highest scores—R@0.5 (70.36), R@0.7 (55.25), mAP@0.5 (69.53), mAP@0.75 (49.17), and average mAP (47.94)—exceeding prior methods, while \textbf{\toolnospaceb} also performing strongly without pretraining. Notably, \textbf{\tool} improves R@0.5 by 6.81\% over UVCOM and 5.70\% over TR-DETR, and average mAP by 4.76\% and 4.94\% over the same baselines. In the HD task, \textbf{\toolnospacebpt} and \textbf{\toolnospaceb} achieve mAPs of 42.84 and 42.43, and HIT@1 scores of 70.56 and 68.94, respectively, outperforming other methods. Even models with fewer features (\textbf{\toolnospace} and \textbf{\toolnospacept}) remain competitive, demonstrating the scalability of our approach. In general, improvements ranging from 2. 76\% to 7. 07\% on various metrics highlight the effectiveness of our framework, with the integration of additional features (e.g., BLIP) further enhancing performance in video language understanding.

\begin{table}[!t]
\centering
\caption{Results on Charades-STA test set. Bold represents the best result, and underline represents the 2nd best result.}
\small
\label{tab:result_charades-sta}
    \vspace{-5pt}
 \begin{tabular}{@{}lcccc@{}}
    \toprule
    Method         & R@0.3 & R@0.5 & R@0.7 &mIoU  \\ \midrule
    2D-TAN (AAAI, 2020)~\cite{zhang2020learning} & 58.76 & 46.02 & 27.5 & 41.25 \\
    VSLNet (ACL, 2020)~\cite{zhang2020span}         & 60.30  & 42.69 & 24.14 & 41.58 \\
    Moment-DETR (NIPS, 2021)~\cite{lei2021detecting}    & 65.83 & 52.07 & 30.59 & 45.54 \\
    QD-DETR (CVPR, 2023)~\cite{Moon_2023_CVPR}        & -     & 57.31 & 32.55 &  -     \\
    TR-DETR (AAAI, 2024)~\cite{Sun_Zhou_Chen_Xie_2024}        & -     & 57.61 & 33.52 &  -     \\
    UniVTG (ICCV, 2023)~\cite{lin2023univtg}         & {\bf 70.81 } & \underline{58.01} & \underline{35.65} & \underline{50.10}  \\
    \rowcolor[gray]{.8}
    \textbf{\toolnospace}     & \underline{70.67} & \textbf{ 58.04} & {\bf 36.88} & {\bf 50.20} \\
    \midrule
    UniVTG(pt) (ICCV, 2023)~\cite{lin2023univtg}  & {\bf 72.63} & {\bf 60.19} & {\bf 38.55} & {\bf 52.17} \\
    \rowcolor[gray]{.8}
    \textbf{\toolnospacept} & \underline{72.26}  & \underline{60.11} & \underline{37.80} & \underline{51.44} \\ 
    \midrule
    \rowcolor[gray]{.8}
    \textbf{\toolnospaceb}     & {\bf 71.72} & {\bf 60.30} & {\bf 37.23} & {\bf 51.25} \\
    \rowcolor[gray]{.8}
    \textbf{\toolnospacebpt}     & {\bf 73.33} & {\bf 61.96} & {\bf 41.05} & {\bf 52.94} \\
    \bottomrule
\end{tabular}
\vspace{-18pt}
\end{table}

\begin{table*}[!t]
\caption{Ablation study on different modules and losses on QVHighlights val split. Here fra stands for FRA module, bi stands for Bi-CMF module, bf stans for Blip features, pt stands for pre-train on the synthetic dataset using Blip Backend, and adaptive hard positive / negative ($\mathcal{L}_\text{hdl}$), task-coupled ($\mathcal{L}_\text{tc}$), task-specific ($\mathcal{L}_\text{ts}$), and alignment ($\mathcal{L}_\text{align}$) loss. The effect of different pretraining data is in the bottom block.}
\vspace{-5pt}
\small
\label{tab:effect_modules}
\centering
\begin{tabular}{@{}ccccc|cccc|ccccccc@{}}
\toprule
\multicolumn{5}{c}{\multirow{2}{*}{Modules}} & \multicolumn{4}{c}{\multirow{2}{*}{Losses}} & \multicolumn{5}{c}{MR} & \multicolumn{2}{c}{HD} \\ \cmidrule(l){10-16} 
\multicolumn{5}{c}{} & \multicolumn{4}{c}{} & \multicolumn{2}{c}{R1} & \multicolumn{3}{c}{mAP} & \multicolumn{2}{c}{\textgreater{}=Very Good} \\ \midrule
sl. &fra & bi & bf & pt & $\mathcal{L}_\text{hdl}$ & $\mathcal{L}_\text{tc}$ & $\mathcal{L}_\text{ts}$ & $\mathcal{L}_{\text{align}}$ & @0.5 & @0.7 & @0.5 & @0.75 & Avg & mAP & HIT@1 \\ \midrule
1. & \xmark & \xmark &  \xmark & \xmark & \cmark & \cmark & \cmark & \cmark & 61.42 & 46.77 & 60.82 & 41.36 & 41.28 & 38.08 & 60.45 \\
2. & \xmark & \xmark &  \cmark & \xmark & \cmark & \cmark & \cmark & \cmark & 64.45 & 49.48 & 63.69 & 43.08 & 43.28 & 39.98 & 64.13 \\
3. & \cmark & \cmark &  \xmark & \xmark & \cmark & \cmark & \cmark & \cmark & 66.77 & 51.23 & 65.83 & 45.38 & 45.12 & 40.74 & 66.9 \\
4. & \xmark & \cmark &  \cmark & \xmark & \cmark & \cmark & \cmark & \cmark & 65.42 & 52.84 & 64.89 & 46.67 & 45.69 & 40.75 & 65.55 \\
5. & \cmark & \xmark &  \cmark & \xmark & \cmark & \cmark & \cmark & \cmark & 69.55 & 53.94 & 67.53 & 47.86 & 47.14 & 42.09 & 68.77 \\
6. & \cmark & \cmark &  \cmark & \xmark & \cmark & \cmark & \cmark & \cmark & 70.06 & 55.35 & 68.75 & 49.22 & 48.44 & 42.84 & 70.71 \\
7. & \cmark & \cmark & \cmark & \xmark & \xmark & \xmark & \xmark & \xmark & 69.29 & 53.03 & 68.76 & 47.36 & 47.19 & 41.82 & 68.00 \\
8. & \cmark & \cmark & \cmark & \xmark & \cmark & \xmark & \xmark & \xmark & 70.19 & 54.77 & 68.59 & 49.00 & 48.35 & 42.73 & 69.10 \\
9. & \cmark & \cmark & \cmark & \xmark & \xmark & \cmark & \xmark & \xmark & 69.55 & 54.00 & 68.37 & 47.80 & 47.63 & 41.85 & 69.61 \\
10. & \cmark & \cmark & \cmark & \xmark & \xmark & \xmark & \cmark & \xmark & 69.81 & 54.39 & 69.06 & 49.21 & 48.56 & 42.76 & 69.74 \\
11. & \cmark & \cmark & \cmark & \xmark & \xmark & \xmark & \xmark & \cmark & 69.68 & 54.71 & 67.80 & 47.80 & 46.68 & 41.79 & 68.26 \\ 
12. & \cmark & \cmark & \xmark & \cmark & \cmark & \cmark & \cmark & \cmark &  71.03 & 54.84 & 68.07 & 47.36 & 46.06 & 42.16 & 69.16 \\
13. & \cmark & \cmark &  \cmark & \cmark & \cmark & \cmark & \cmark & \cmark &  \textbf{72.06} & \textbf{57.94} & \textbf{70.38} & \textbf{51.12} & \textbf{49.71} & \textbf{43.12} & \textbf{71.48} \\ \midrule
\multicolumn{9}{c}{No Pretraining} & 66.77 & 51.23 & 65.83 & 45.38 & 45.12 & 40.74 & 66.9 \\
\multicolumn{9}{c}{ASR Pretraining ~\cite{lei2021detecting}} & 67.94 & 51.48 & 65.84 & 44.03 & 43.74 & 40.71 & 67.03 \\ 
\multicolumn{9}{c}{Our BLIP Pretraining} & \textbf{71.03} & \textbf{54.84} & \textbf{68.07} & \textbf{47.36} & \textbf{46.06} & \textbf{42.16} & \textbf{69.16} \\ 
\bottomrule
\end{tabular}
\end{table*}

\noindent \textbf{Performance on Charades-STA:}
Our models (\textbf{\tool}, \textbf{\toolpt}, \textbf{\toolb}, and \textbf{\toolbpt})  achieve strong results on Charades-STA (Table~\ref{tab:result_charades-sta}). Without pretraining, \textbf{\tool} attains state-of-the-art (SOTA) in three metrics: R@0.5 (58.04 vs. UniVTG’s 58.01), R@0.7 (36.88 vs. 35.65), \& mIoU (50.20 vs. 50.10), while lagging slightly in R@0.3 (70.67 vs. 70.81). Pretrained \textbf{\toolpt} performs competitively, trailing UniVTG (pt) by 0.8\% across metrics.
 \textbf{\toolb~}, incorporating BLIP features, outperforms UniVTG in R@0.5 (60.30 vs. 58.01) and mIoU (51.25 vs. 50.10) without pretraining. Pre-trained \textbf{\toolbpt} achieves SOTA in all metrics: R@0.3 (73.33), R@0.5 (61.96), R@0.7 (41.05), \& mIoU (52.94), surpassing UniVTG (pt) by 0.70–2.50\%. These results underscore the effectiveness of integrating BLIP features and pretraining, establishing new benchmarks across all metrics.

\begin{figure*}[t]
    \centering
    \includegraphics[width=\textwidth]{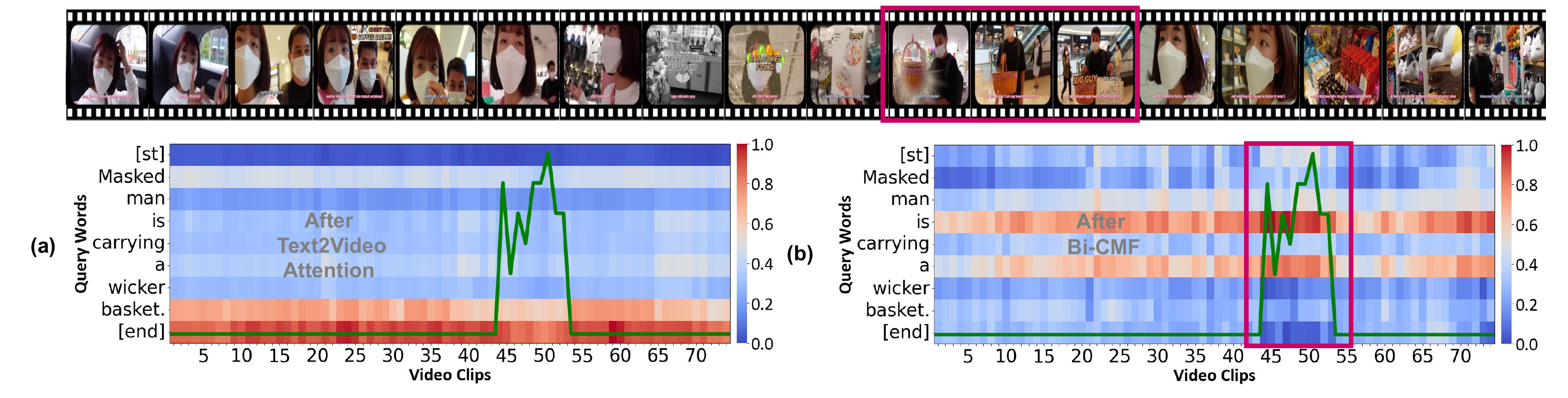}
     \vspace{-24pt}
    \caption{(a) and (b) show video-query correspondence maps: (a) after text-to-video (t2v) attention and (b) after the Bi-CMF layer. The green line represents the ground truth saliency scores. Bi-CMF attends to the correct video region better than t2v (highlighted in the magenta box). The word `Is' asserts that `a' refers to one basket, unlike `is not'.} 
    \vspace{-4pt}
    \Description{Effect of Bi-CMF}
    \label{fig:effect_bicmf}
\end{figure*}

\noindent \textbf{Perfomance in TVSum:}
\textbf{\toolnospace} achieves strong results on TVSum, securing state-of-the-art performance in 5 out of 10 domains and matching the best overall average (87.9\%) while closely trailing TR-DETR (88.1\%). It outperforms TR-DETR in key domains such as GA, MS, PR, VT, and BK, and remains competitive in others. Compared to UniVTG, both \textbf{\toolnospace} and \textbf{\toolnospacept} exhibit substantial gains, with \textbf{\toolnospace} exceeding UniVTG by 6.9\% in overall average. The pretrained variant \textbf{\toolnospacept} further improves performance, achieving state-of-the-art results in 7 domains and outperforming UniVTG (pt) by 3.3\%. Variants incorporating BLIP features (\textbf{\toolnospaceb}, \textbf{\toolnospacebpt}) also deliver competitive to superior results, especially in VU, GA, BK, and DS. These findings underscore the effectiveness and scalability of \textbf{\tool} in various HD scenarios.

\begin{figure*}[t]
    \centering
    \includegraphics[width=\textwidth]{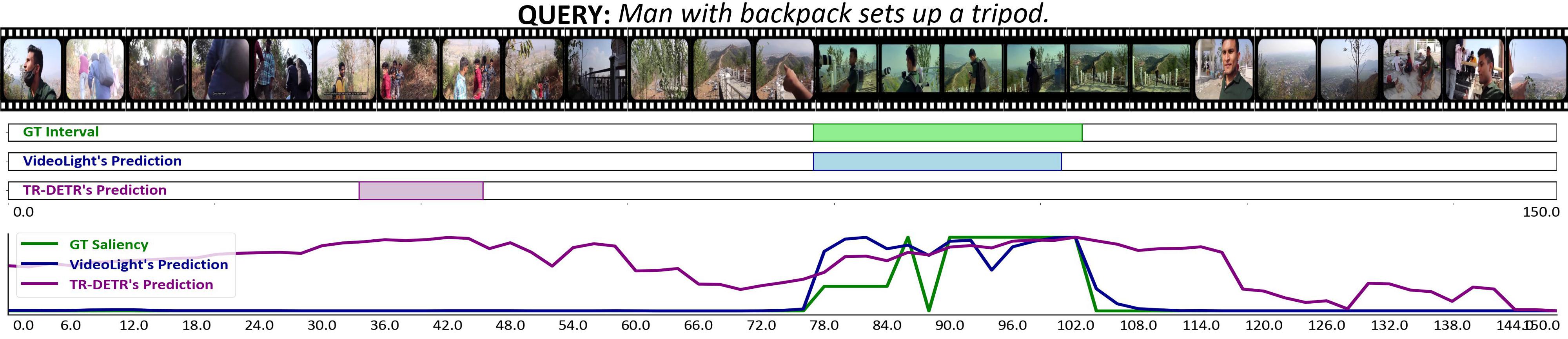}
    \caption{Qualitative results. It demonstrates \textbf{\tool} outperformed TR-DETR~\cite{Sun_Zhou_Chen_Xie_2024} in both MR and HD.} 
    \vspace{-4pt}
    \Description{Qualitative results.}
    \label{fig:qualitative_result}
\end{figure*}

In summary, \textbf{\toolnospace} not only matches but often exceeds the performance of other cutting-edge methods, demonstrating its effectiveness in joint video highlight detection \& moment retrieval. Beyond these primary benchmarks, we demonstrate the model's broader generalization capabilities with additional successful evaluations on the TACoS\cite{regneri-etal-2013-grounding} and NLQ (Ego4D)\cite{Grauman_2022_CVPR} datasets, as detailed in Appendix~\ref{sec:additional_experiments}. Along with the quantitative results, Figure~\ref{fig:qualitative_result} presents qualitative results in QVHighlights, where \tool accurately localizes the 'tripod setup' activity, unlike TR-DETR, demonstrating superior handling of complex queries. Additional examples are provided in Appendix~\ref{sec:qualitative_resuts}.

\subsection{Ablation Studies}
\label{sec:ablation}

To comprehend module impacts, we present our model ablation on QVHighlights \emph{val} split in Table~\ref{tab:effect_modules}.


\noindent\textbf{Effect of FRA:}  
As shown in Table~\ref{tab:effect_modules} (rows 2 vs. 5) and qualitatively in Figure~\ref{fig:fra}, adding the FRA module (with Bi-CMF disabled) yields an average performance improvement of 7.93\% (ranging from 5.28\% to 11.09\%). Table~\ref{tab:fra_on_diff_methods} in Appendix~\ref{sec:additional_ablation_fra} further demonstrates that incorporating FRA consistently benefits various baselines, with larger gains observed in weaker models (e.g., Moment-DETR) and modest improvements in stronger ones (e.g., QD-DETR and TR-DETR).


\noindent\textbf{Effect of Bi-CMF:}  
Table~\ref{tab:effect_modules} (rows 2 vs. 4) shows that our Bi-CMF module provides an average improvement of 4.03\%, with a notable increase in mAP@0.75 (8.33\%). Feature heatmap visualizations (Figure~\ref{fig:effect_bicmf}) indicate that Bi-CMF produces a sparser and more discriminative attention spectrum compared to both the baseline and Uni-CMF. This is corroborated by Table~\ref{tab:unicmf_vs_bicmf} in Appendix~\ref{sec:bicmf_motivation}, where Bi-CMF outperforms Uni-CMF across all metrics, especially in HIT@1 (+1.94) and R1@0.75 (+1.41).


\noindent\textbf{Effect of New Loss Functions:}  
Rows 6 to 11 in Table~\ref{tab:effect_modules} illustrate that each proposed loss - adaptive hard positive / negative ($\mathcal{L}_\text{hdl}$), task-coupled ($\mathcal{L}_\text{tc}$), task-specific ($\mathcal{L}_\text{ts}$), and alignment loss ($\mathcal{L}_\text{align}$) - independently improves MR and HD performance. Their combined use (Row 6) achieves the best overall results, highlighting their synergistic effect.


\noindent\textbf{Effect of BLIP-2 Features and Pretraining:}  
Pretraining further improves performance, as evidenced by the difference between the 6th and 11th rows in the upper block of Table~\ref{tab:effect_modules}. Table~\ref{tab:effect_feature} in Appendix~\ref{sec:effect_features} shows that replacing CLIP with BLIP-2 features yields gains, with the best results achieved by integrating SlowFast, CLIP, and BLIP-2. The bottom block of Table~\ref{tab:effect_modules} confirms that BLIP pretraining outperforms ASR pretraining, with gains between 3.18\% and 7.57\%.

\section{Limitation and Conclusion}\label{sec:limitation_conclution}
We introduce \textbf{\toolnospace}, a novel framework that jointly addresses video highlight detection (HD) and moment retrieval (MR) through innovative cross-modal and cross-task interactions. \textbf{\toolnospace} achieves state-of-the-art performance on benchmark datasets including QVHighlights, TVSum, and Charades-STA by integrating several key components: the Feature Refinement and Alignment (FRA) module for effective local and global feature alignment; the Bi-Directional Cross-Modal Fusion (Bi-CMF) network for enhanced query-aware representations; and the Unidirectional Joint-Task Feedback Mechanism (Uni-JFM) to optimize both task-specific and cross-task learning. We further leverage LVLM features (e.g., BLIP-2) to improve temporal awareness and semantic alignment, and employ intelligent synthetic data generation and pre-training to bolster performance and robustness. Comprehensive evaluations and ablation studies confirm the superiority of \textbf{\toolnospace} over previous baselines. Future work may explore advanced multimodal fusion, improved feature alignment, and broader real-world applications, as well as further investigation into LVLMs for moment retrieval.


\noindent\textbf{Limitation:}  
While \textbf{\tool} introduces architectural complexity, our ablation results (Table~\ref{tab:effect_modules}) confirm that each module contributes meaningfully, with the full model yielding synergistic gains. Weakly supervised pretraining may introduce caption bias, but this is mitigated via fine-tuning on human-annotated data and alignment loss (Appendix~\ref{sec:bias_filtering}). Additionally, our reliance on pre-trained models for captioning and feature extraction incurs computational overhead and dependence on external resources, potentially limiting scalability. The performance of the Bi-CMF module also hinges on the quality of input features and the effectiveness of the attention mechanisms, which can vary with the complexity and diversity of video content. Addressing these challenges through further research is essential to fully realize the potential of our approach in real-world settings.

\bibliographystyle{ACM-Reference-Format}
\bibliography{references}

\appendix

\section{Appendix}
\label{sec:appendix}

\subsection{Dataset statistics}
\label{sec:dataset_statistics}
Table~\ref{tab:datasets} provides a comparison of three datasets utilized in a study, describing the different attributes of each. The QVHighlights dataset includes vlog and news content, with 10,300 annotations and 12,500 videos. It supports tasks such as Moment Retrieval (MR) and Highlight Detection (HD) and has been utilized in pre-training. We have generated 187682 synthetic data from videos of this dataset using the approach described in Algorithm~\ref{alg:data_gen}. The Charades-STA dataset, which focuses on activity-related content, comprises 16,100 annotations and 6,700 videos, specifically used for Moment Retrieval and has also been employed in pre-training. We have generated 23,193 synthetic data samples from this dataset. Lastly, the TVSum dataset, based on web content, is notably smaller, with 50 annotations and 50 videos, used exclusively for Highlight Detection. It has 10 domains, VT, VU, GA, MS, PK, PR, FM, BK, BT, and DS each containing 5 videos. Unlike the other datasets, it has not been used in pre-training and does not include synthetic data.

\begin{table*}[!t]
\caption{Comparison of datasets used in this study.}
\vspace{-5pt}
\centering
\small
\label{tab:datasets}
\begin{tabular}{@{}lcccccc@{}}
\toprule
Dataset & Domain & Annotations & Videos & Task & Used in pt & Synthetic data  \\ \midrule
QVHighlights & Vlog / News & 10.3K & 12.5K & MR, HD & \cmark & 187682 \\
Charades-STA & Activity & 16.1K & 6.7K & MR & \cmark & 23193 \\
TVSum & Web & 50 & 50 & HD &  & \\ \bottomrule
\end{tabular}
\vspace{-5pt}
\end{table*}

\begin{algorithm}[!t]
\caption{Synthetic Data Generation Process}
\label{alg:data_gen}
\begin{algorithmic}[1]
    \REQUIRE Input video $\mathcal{V}$ with duration $T$
    \ENSURE Synthetic dataset $\mathcal{D}_{\text{synthetic}}$

    \STATE Divide the video $\mathcal{V}$ into $n = \lceil T / 10 \rceil$ non-overlapping intervals $\{I_1, I_2, \dots, I_n\}$, where each interval $I_i$ corresponds to a 10-second segment of $\mathcal{V}$.
    
    \FOR{each interval $I_i$} 
        \STATE Select a representative frame $f_i$ from $I_i$ (e.g., the middle frame or one sampled by a heuristic).
        \STATE Use the BLIP-2 model $\mathcal{M}_{\text{BLIP}}$ to generate a caption $c_i = \mathcal{M}_{\text{BLIP}}(f_i)$ describing the content of $f_i$.
    \ENDFOR

    \FOR{each interval $I_i$}
        \FOR{each frame $f_{ij} \in I_i$}
            \STATE Compute the cosine similarity $\text{Sim}(c_i, f_{ij})$ between the caption $c_i$ and the frame $f_{ij}$ using their feature representations $\phi(c_i)$ and $\phi(f_{ij})$.
        \ENDFOR
        \STATE Use $s_i = \text{Sim}(c_i, f_{ij})$ for each video frame $f_{ij}$ as frame-wise highlight scores for interval $I_i$.
    \ENDFOR

    \STATE Construct a synthetic dataset $\mathcal{D}_{\text{synthetic}} = \{(c_i, I_i, s_i) \mid i \in [1, n]\}$, where $c_i$ is the generated caption $I_i$ is the corresponding interval and $s_i$ is the saliency score.

    \STATE Use $\mathcal{D}_{\text{synthetic}}$ to train the target model for highlight detection or related tasks.
\end{algorithmic}
\end{algorithm}

\begin{table}[!t]
\vspace{-6pt}
\caption{Effect of Linear projection vs Convolutional projection in FRA using CLIP, BLIP, and SlowFast features}
\vspace{-5pt}
\label{tab:effect_projection_type}
\small
\begin{tabular}{@{}cccccc@{}}
\toprule
\multirow{2}{*}{Projection type} & \multicolumn{3}{c}{MR} & \multicolumn{2}{c}{HD} \\ \cmidrule(l){2-6} 
 & R1@0.5 & R1@0.75 & mAP@Avg & mAP & HIT@1 \\ \midrule
Linear & 69.81 & 54.52 & 48.15 & 42.57 & 69.68 \\
Convolutional & \textbf{70.06} & \textbf{55.35} & \textbf{48.44} & \textbf{42.84} & \textbf{70.71} \\ \bottomrule
\end{tabular}
\vspace{-5pt}
\end{table}

\begin{table}[!t]
\caption{Effect of FRA on different methods on QVHighlights \textbf{val} set.  \(\dag\) represents the use of the FRA module.}
\vspace{-5pt}
\label{tab:fra_on_diff_methods}
\small
\resizebox{.49\textwidth}{!}{
\begin{tabular}{@{}lccccc@{}}
\toprule
\multicolumn{1}{c}{\multirow{2}{*}{Method}} & \multicolumn{3}{c}{MR} & \multicolumn{2}{c}{HD} \\ \cmidrule(l){2-6} 
\multicolumn{1}{c}{} & R1@0.5 & R1@0.75 & mAP@Avg & mAP & HIT@1 \\ \midrule
Moment-DETR~\cite{lei2021detecting} & 53.94 & 34.84 & 32.2 & 35.36 & 55.55 \\
Moment-DETR \(\dag\) & \textbf{61.48} & \textbf{40.26} & \textbf{35.17} & \textbf{38.88} & \textbf{63.16} \\ \midrule
QD-DETR~\cite{Moon_2023_CVPR} & 62.68 & 46.66 & 41.22 & 39.13 & 63.03 \\
QD-DETR \(\dag\) & \textbf{63.81} & \textbf{46.84} & \textbf{41.71} & \textbf{39.77} & \textbf{63.87} \\ \midrule
TR-DETR~\cite{Sun_Zhou_Chen_Xie_2024} & 67.10 & 51.48 & 45.09 & 40.55 & 64.77 \\
TR-DETR \(\dag\) & \textbf{67.81} & \textbf{51.68} & \textbf{45.19} & \textbf{41.37} & \textbf{67.03} \\ \bottomrule
\end{tabular}
}
\vspace{-5pt}
\end{table}

\begin{table}[!th]
\caption{CMFs in \tool on QVHighlights (\textbf{val})}
\vspace{-5pt}
\label{tab:unicmf_vs_bicmf}
\resizebox{.47\textwidth}{!}{
\begin{tabular}{@{}cccccc@{}}
\toprule
\multirow{2}{*}{Cross-Attention Type} & \multicolumn{3}{c}{MR} & \multicolumn{2}{c}{HD} \\ \cmidrule(l){2-6} 
 & R1@0.5 & R1@0.75 & mAP@Avg & mAP & HIT@1 \\ \midrule
Bi-CMF & \textbf{70.06} & \textbf{55.35} & \textbf{48.44} & \textbf{42.84} & \textbf{70.71} \\
Uni-CMF & 69.55 & 53.94 & 47.14 & 42.09 & 68.77 \\ \bottomrule
\end{tabular}
\vspace{-5pt}
}
\end{table}

\subsection{Additional Experiment on dataset form different domains}
\label{sec:additional_experiments}

\begin{table*}[!th]
\caption{Performance Comparison on TACoS and NLQ (Ego4D) using Clip and SlowFast visual features.}
\vspace{-5pt}
\label{tab:result_tacos_nlq}
\begin{tabular}{@{}l|cccc|cccc@{}}
\toprule
\multirow{2}{*}{Model}          & \multicolumn{4}{c|}{TACos}    & \multicolumn{4}{c}{NLQ (Ego4d)} \\ \cmidrule(l){2-9} 
                                & R@0.3 & R@0.5 & R@0.7 & mIoU  & R@0.3  & R@0.5  & R@0.7  & mIoU \\ \midrule
2D TAN (AAAI, 2020)~\cite{zhang2020learning}    & 40.01 & 27.99 & 12.92 & 27.22 & 4.33   & 1.83   & 0.6    & 3.39 \\
VSLNet (ACL, 2020)~\cite{zhang2020span}    & 35.54 & 23.54 & 13.15 & 24.99 & 4.54   & 2.4    & 1.01   & 3.54 \\
Momoent-DETR (NIPS, 2021)~\cite{lei2021detecting} & 37.97 & 24.67 & 11.97 & 25.49 & 4.34   & 1.81   & 0.65   & 3.53 \\
UniVTG (ICCV, 2023)~\cite{lin2023univtg}       & 51.44 & 34.97 & 17.35 & 33.6  & 7.28   & 3.95   & 1.32   & 4.91 \\
\rowcolor[gray]{.8}
\toolnospace                     & \textbf{54.29} & \textbf{40.61} & \textbf{22.67} & \textbf{37.18} & \textbf{7.56}   & \textbf{4.28}   & \textbf{1.65}   & \textbf{5.64} \\ \bottomrule
\end{tabular}
\vspace{-5pt}
\end{table*}

To assess the generalization ability of VideoLights across diverse domains and video characteristics, we conducted evaluations on the TACoS~\cite{regneri-etal-2013-grounding} and NLQ (Ego4D)~\cite{Grauman_2022_CVPR} datasets, which feature variable-length cooking videos and egocentric videos, respectively. VideoLights achieved superior performance compared to existing baselines on both datasets, demonstrating its robustness and adaptability (Table~\ref{tab:result_tacos_nlq}). These results indicate that VideoLights effectively handles videos with varying lengths and complexities, affirming its generalization capabilities across different domains.

\subsection{Additional ablation on FRA}
\label{sec:additional_ablation_fra}
Additional ablations evaluated convolutional projection vs. linear projection within the FRA module (Table~\ref{tab:effect_projection_type}), demonstrating consistent improvements (0.35\% to 1.52\%) across all metrics. Qualitative results (Figure~\ref{fig:fra}) further illustrate that convolutional projection achieves superior local (word-level) alignment between video segments and textual elements compared to linear projection.

The results in Table~\ref{tab:fra_on_diff_methods} clearly demonstrate that integrating the FRA module consistently improves performance across different methods on the QVHighlights validation set. In particular, the Moment-DETR baseline exhibits substantial gains across all metrics, indicating that the FRA module is especially effective in enhancing weaker models. Even for stronger models like QD-DETR and TR-DETR, the addition of FRA yields measurable improvements in both moment retrieval and highlight detection tasks, underscoring its role in refining feature alignment and boosting overall performance.

\begin{table}[!t]
\caption{Effect of integrating features from different VLM's on \textbf{\tool} on QVHighlights \textbf{val} set. Here SF stands for SlowFast, C stands for CLIP, and B stands for BLIP-2.}
\vspace{-5pt}
\label{tab:effect_feature}
\small
\resizebox{.49\textwidth}{!}{
\begin{tabular}{@{}cccccc@{}}
\toprule
\multirow{2}{*}{Feature type} & \multicolumn{3}{c}{MR} & \multicolumn{2}{c}{HD} \\ \cmidrule(l){2-6} 
 & R1@0.5 & R1@0.75 & mAP@Avg & mAP & HIT@1 \\ \midrule
SF + C & 66.77 & 51.23 & 45.12 & 40.74 & 66.9 \\
SF + B & 69.23 & 53.42 & 46.86 & 42.20 & 69.68 \\
SF + C + B & \textbf{70.06} & \textbf{55.35} & \textbf{48.44} & \textbf{42.84} & \textbf{70.71} \\ \bottomrule
\end{tabular}
}
\vspace{-5pt}
\end{table}

\begin{table}[!t]
\caption{Comparison with BLIP-2-enhanced models. Here \emph{-B} indicates usage of BLIP-2 features}
\vspace{-5pt}
\label{tab:effect_blip_other_methods}
\small
\resizebox{.49\textwidth}{!}{
\begin{tabular}{@{}lccccc@{}}
\toprule
\multirow{2}{*}{Method} & \multicolumn{3}{c}{MR} & \multicolumn{2}{c}{HD} \\ \cmidrule(l){2-6} 
 & R1@0.5 & R1@0.75 & mAP@Avg & mAP & HIT@1 \\ \midrule
 Moment-DETR-B & 58.65 & 38.97 & 35.65 & 38.96 & 61.94 \\
 QD-DETR-B & 65.94 & 50.13 & 43.75 & 40.9 & 65.35 \\
 TaskWeave-B & 67.68 & 51.74 & 46.52 & 40.19 & 64.71 \\
 TR-DETR-B & \textbf{71.61} & 54.77 & 47.4 & 42.9 & 69.81 \\
 \toolb & 70.06 & \textbf{55.35} & \textbf{48.44} & \textbf{42.84} & \textbf{70.71} \\ \bottomrule
\end{tabular}
}
\vspace{-5pt}
\end{table}

\subsection{How Bi-CMF is different from existing works}
\label{sec:bicmf_motivation}

Our Bi-CMF module implements a novel three-stage sequential bidirectional fusion, in contrast to conventional parallel co-attention frameworks ~\cite{tan-bansal-2019-lxmert, li-jiang-2020-two,badamdorj2021joint} or simpler task-specific schemes ~\cite{Yuan2019aaaimoment}. It first produces text-conditioned video tokens and video-conditioned text tokens, then—crucially—enables these refined representations to mutually attend in a final fusion step. This hierarchical decomposition enhances semantic disambiguation for highlight detection and moment retrieval, while its modular design promotes stable, targeted learning of complex video–text relationships. Figure \ref{fig:effect_bicmf} illustrates how Bi-CMF captures video-query correspondence more accurately than standard text-to-video attention, and Table~\ref{tab:unicmf_vs_bicmf} shows its effectiveness over t2v cross attention.

\subsection{Effect of features from different encoders}
\label{sec:effect_features}
Integrating BLIP-2 (B) with SlowFast (SF) features substantially improves performance over the SF+CLIP (C) baseline across all metrics (Table~\ref{tab:effect_feature}), with gains of +2.46\% (R1@0.5), +2.19\% (R1@0.75), and +2.80\% (HIT@1). The combined SF+C+B configuration achieves the best results, outperforming SF+C by +3.29\% (R1@0.5), +4.12\% (R1@0.75), and +3.81\% (HIT@1), and surpassing SF+B in mAP@Avg (+1.58\%) and mAP (+0.64\%). This demonstrates the complementary strengths of CLIP and BLIP-2 features, where their joint integration maximizes moment retrieval (MR) and highlight detection (HD) accuracy, suggesting synergistic benefits from multi-modal visual-language representations.

We also conducted controlled experiments integrating BLIP-2 into existing methods. As shown in Table~\ref{tab:effect_blip_other_methods}, BLIP-2 improves performance across baselines; however, \textbf{\toolb} still outperforms all others on most metrics, indicating the strength of our model design beyond feature selection. We note that not all baselines could be evaluated due to time constraints.

\subsection{Removing biases and noises introduced in pretraining}
\label{sec:bias_filtering}

\begin{table}[!t]
\caption{Comparing different variants of VideoLights. Here '-ZS' means Zero shot, '-B' means incorporating BLIP-2 features and '-PT' means pretrained.}
\vspace{-5pt}
\label{tab:bias_filtering}
\small
\begin{tabular}{@{}lccccc@{}}
\toprule
\multirow{2}{*}{Method} & \multicolumn{3}{c}{MR} & \multicolumn{2}{c}{HD} \\ \cmidrule(l){2-6} 
 & R1@0.5 & R1@0.75 & mAP@Avg & mAP & HIT@1 \\ \midrule
 \toolzs & 22.00 & 6.77 & 6.56 & 37.97 & 58.97 \\
 \tool & 66.77 & 51.23 & 45.12 & 40.74 & 66.9 \\
 \toolpt & \textbf{71.03} & \textbf{54.84} & \textbf{46.06} & \textbf{42.16} & \textbf{69.16} \\ \midrule
 \toolbzs & 24.26 & 7.23 & 6.69 & 40.70 & 66.52 \\
 \toolb & 70.06 & 55.35 & 48.44 & 42.84 & 70.71 \\
 \toolbpt & \textbf{72.06} & \textbf{57.94} & \textbf{49.71} & \textbf{43.12} & \textbf{71.48} \\ \bottomrule
\end{tabular}
\vspace{-5pt}
\end{table}

Weakly supervised pretraining using synthetically generated captions can introduce biases or inaccuracies, particularly when such captions do not accurately reflect the underlying video content. However, our approach mitigates these potential issues through a subsequent fine-tuning stage on human-annotated datasets, which are typically of higher fidelity and less prone to the same biases.
Recent studies \cite{wang2023overwriting, wu-etal-2022-noisytune} demonstrate that fine-tuning on clean, task-specific data can effectively override biases acquired during pretraining, thus enhancing the model's generalization capability. In our case, the contrastive and alignment losses employed during fine-tuning further enforce consistency between query semantics and video content, thereby refining the model's attention towards relevant temporal regions. Our findings are consistent with these observations: the transition from pre-training to supervised fine-tuning yields substantial performance gains, as evidenced by improved results on unseen test data (see Table~\ref{tab:bias_filtering}). Specifically, while the zero-shot variant \textbf{\toolzs} and \textbf{\toolbzs} underperforms due to noise in the pretraining data, the finetuned version \textbf{\toolpt} and \textbf{\toolbpt} significantly outperforms models trained from scratch, highlighting the utility of this two-stage learning paradigm.

\begin{table*}[!th]
\centering
\caption{Experiment-specific hyperparameters. Features: I3D, SlowFast (SF), CLIP (C), and BLIP-2 (B). VF: visual features, TF: text features. Coefficients: symmetric alignment loss ($\lambda_\text{al}$), task coupled loss ($\lambda_\text{tc}$), hard positive/negative loss ($\lambda_\text{hdl}$), and cosine similarity loss ($\lambda_\text{ts}$).}
\vspace{-5pt}
\label{tab:exp_details}
\small
\begin{tabular}{@{}lcccccccccc@{}}
\toprule
Dataset & Model & VF & TF & Epoch & lr & Bs & $\lambda_\text{al}$ & $\lambda_\text{tc}$ & $\lambda_\text{hdl}$ & $\lambda_\text{ts}$ \\ \midrule
\multicolumn{1}{l}{\multirow{4}{*}{QVHighlights}} & \toolnospace & SF+C & C & 200 & 1e-04 & 32 & 0.01 & 1 & 10 & 1 \\
\multicolumn{1}{l}{} & \toolnospacept & SF+C & C & 200 & 1e-04 & 32 & 0.01 & 1 & 10 & 1 \\
\multicolumn{1}{l}{} & \toolnospaceb & SF+C+B & C+B & 200 & 1e-04 & 32 & 0.2 & 1 & 10 & 1 \\
\multicolumn{1}{l}{} & \toolnospacebpt & SF+C+B & C+B & 200 & 1e-04 & 32 & 0.2 & 1 & 10 & 1 \\\midrule
\multicolumn{1}{l}{\multirow{4}{*}{Charades-STA}} & \toolnospace & SF+C & C & 100 & 1e-04 & 32 & 0.3 & 1 & 10 & 1 \\
\multicolumn{1}{l}{} & \toolnospacept & SF+C & C & 100 & 1e-04 & 32 & 0.3 & 1 & 10 & 1 \\
\multicolumn{1}{l}{} & \toolnospaceb & SF+C+B & C+B & 100 & 1e-04 & 32 & 0.3 & 1 & 10 & 1 \\
\multicolumn{1}{l}{} & \toolnospacebpt & SF+C+B & C+B & 100 & 1e-04 & 32 & 0.3 & 1 & 10 & 1\\ 
\midrule
\multicolumn{1}{l}{\multirow{1}{*}{TACos}} & \toolnospace & SF+C & C & 150 & 2e-04 & 32 & 0.002 & 1 & 1 & 1 \\
\midrule
\multicolumn{1}{l}{\multirow{1}{*}{NLQ (Ego4d)}} & \toolnospace & SF+C & C & 150 & 2e-04 & 32 & 0.002 & 1 & 1 & 1 \\
\bottomrule
\end{tabular}
\vspace{-5pt}
\end{table*}

\balance

\subsection{Qualitative results}
\label{sec:qualitative_resuts}

\begin{figure*}[t]
    \centering
    \vspace{-10pt}
    \subfloat[][]{
        \includegraphics[width=\textwidth]{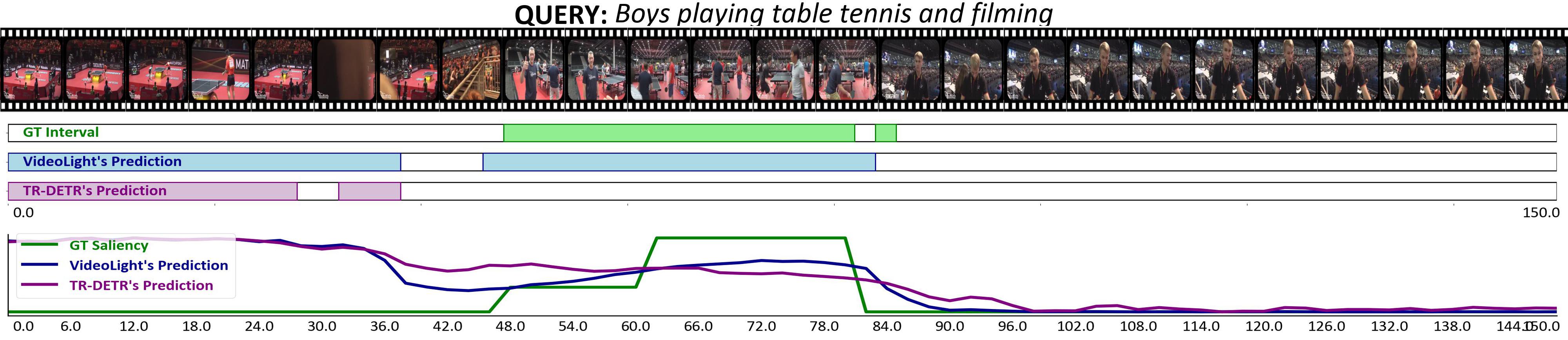}
    }
    \vspace{-10pt}
    \qquad
    \subfloat[][]{
        \includegraphics[width=\textwidth]{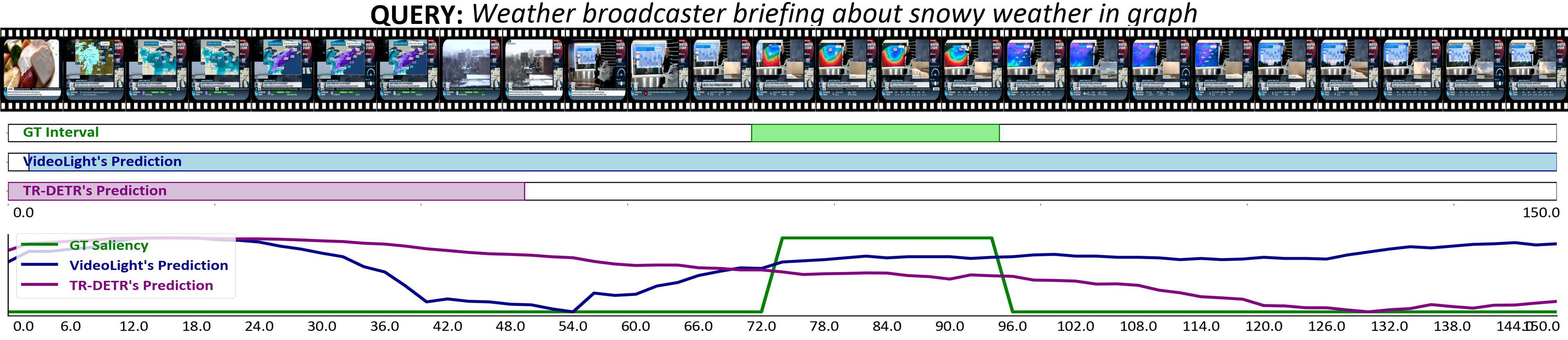}
    }
    \vspace{-10pt}
    \qquad
    \subfloat[][]{
        \includegraphics[width=.45\textwidth]{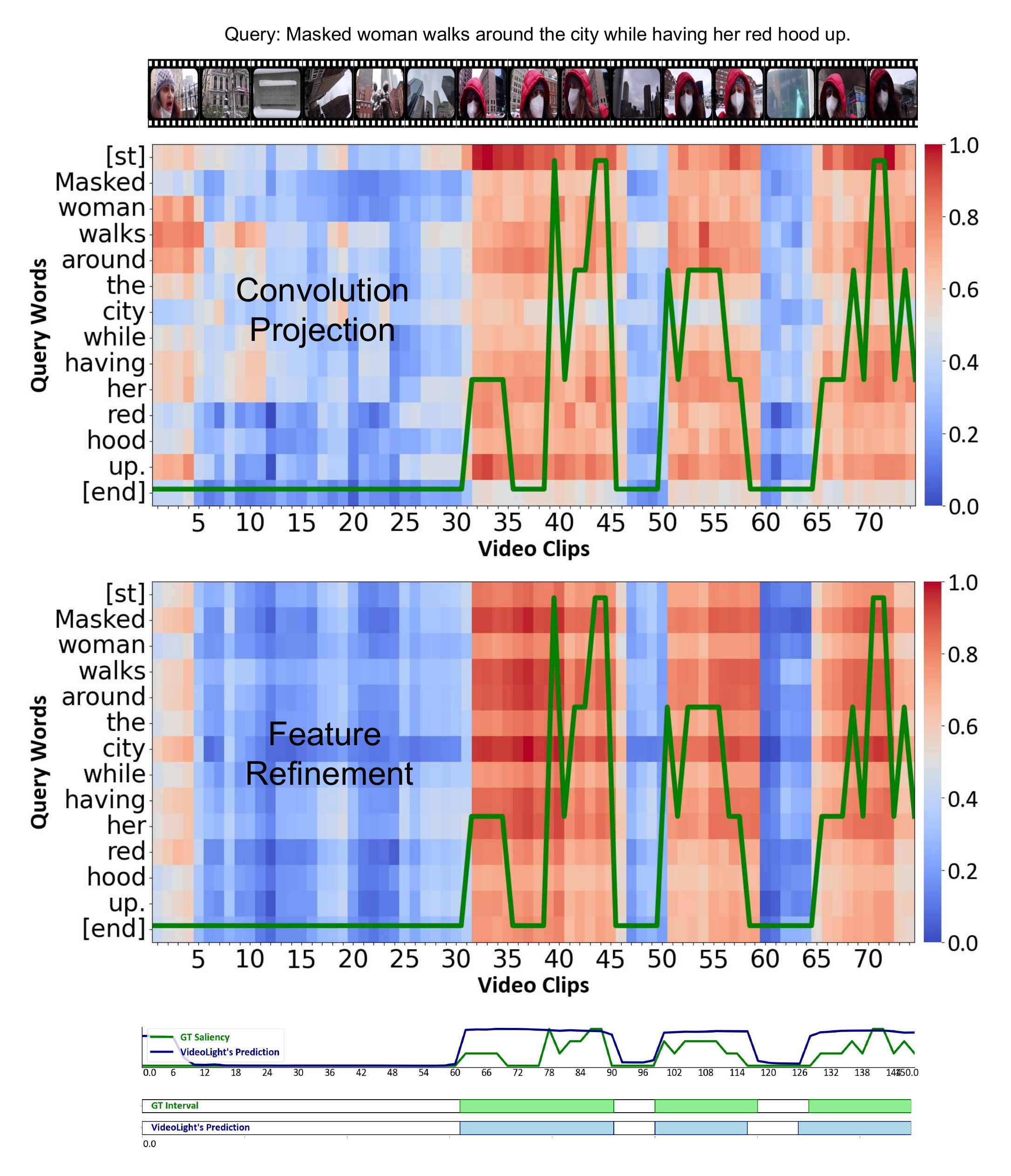}
    }
    \qquad
    \subfloat[][]{
        \includegraphics[width=.45\textwidth]{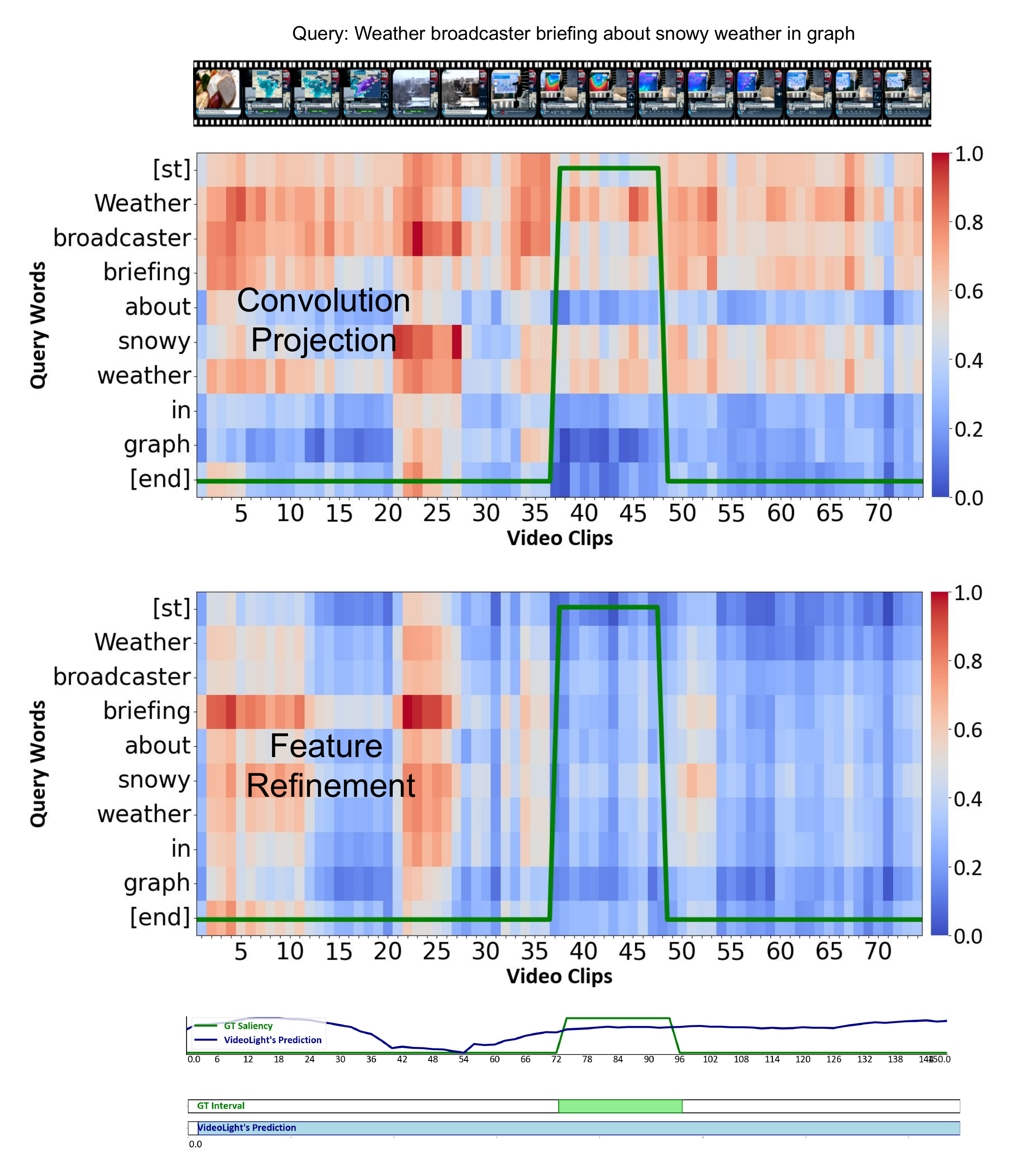}
    }
    \vspace{-10pt}
    \caption{
        Qualitative results: (a) Both \tool and TR-DETR underperform yet mispredicted clips remain query-related; (b) minimal frame changes hinder moment detection; (c) effective FRA alignment yields accurate HD/MR predictions (green: GT, blue: \tool); (d) poor FRA alignment degrades HD/MR performance (green: GT, blue: \tool).
    }
    \Description{Qualitative results.}
    \label{fig:qualitative_result_2}
\vspace{-15pt}
\end{figure*}

Figure~\ref{fig:qualitative_result_2} illustrates \textbf{\toolnospace}'s performance under various conditions using examples from the QVHighligts validation set. In Figure~\ref{fig:qualitative_result_2}~(a), although both \textbf{\tool} and TR-DETR fall short of the ground truth, the mispredicted clips remain semantically related to the query. In contrast, Figure~\ref{fig:qualitative_result_2}~(b) shows that when consecutive frames exhibit little change, the model fails to properly detect key moments. Furthermore, Figure~\ref{fig:qualitative_result_2}~(c) demonstrates that when the FRA effectively aligns video and query features, \textbf{\tool} produces predictions closely matching the ground truth for both highlight detection (HD) and moment retrieval (MR)—as indicated by the green plots (ground truth) versus the blue predictions. Conversely, Figure~\ref{fig:qualitative_result_2}~(d) reveals that poor alignment leads to significant deviations in MR and HD predictions. These qualitative observations underscore the critical role of precise feature alignment in achieving accurate video moment retrieval and highlight detection.

\subsection{Reproducibility Statement}
\label{sec:reproducibility_statement}
To ensure the reproducibility of our experimental results, we provide comprehensive details of our implementation. The core hyperparameters and environmental settings used across all experiments are thoroughly documented in Section~\ref{sec:implementation_details}. For specific experiments that required parameter tuning, we present a detailed breakdown in Table~\ref{tab:exp_details}, which includes the optimal hyperparameter configurations for each dataset and evaluation scenario. This includes learning rates, batch sizes, and model-specific parameters that were determined through empirical validation. The complete source code, including pre-processing scripts, model architectures, training pipelines, and evaluation protocols, along with detailed instructions for environment setup and data preparation, is available in the link in the abstract. We shall provide model checkpoints and experiment logs to ensure reproducibility.




\end{document}